\documentclass{article} 
\usepackage{amsmath}
\usepackage{amssymb}
\usepackage{amsfonts}
\usepackage[margin=2.5cm]{geometry}
\usepackage{subcaption}

\usepackage{cleveref}
\newcommand{\csubref}[1]{(\subref{#1})}
\usepackage{tikz}
\usepackage[edges]{forest}
\usepackage{comment}
\usepackage{multirow}

\usepackage{biblatex} %Imports biblatex package
\usepackage{listings}
\renewcommand{\Re}{\mathbb{R}}

\definecolor{jon}{rgb}{0.79,0.25,0.33}
\definecolor{ken}{rgb}{0.94,0.88,0.18}
\newcommand{\jon}[1]{{\color{jon}{\bf Jon says:} #1}}
\newcommand{\ken}[1]{{\color{ken}{\bf Ken says:} #1}}
\renewcommand{\jon}[1]{}
\renewcommand{\ken}[1]{}

\title{Statistical investigations into the geometry and homology of random programs}
\author{Jon Sporring \& Ken Friis Larsen\\
University of Copenhagen, Department of Computer Science, Copenhagen, Denmark}

\begin{document}
\maketitle
\begin{abstract}
AI-supported programming has taken giant leaps with tools such as Meta's Llama and openAI’s chatGPT. These are examples of stochastic sources of programs and have already greatly influenced how we produce code and teach programming. If we consider input to such models as a stochastic source, a natural question is, what is the relation between the input and the output distributions, between the chatGPT prompt and the resulting program?

In this paper, we will show how the relation between random Python programs generated from chatGPT can be described geometrically and topologically using Tree-edit distances between the program’s syntax trees and without explicit modeling of the underlying space. A popular approach to studying high-dimensional samples in a metric space is to use low-dimensional embedding using, e.g., multidimensional scaling. Such methods imply errors depending on the data and dimension of the embedding space. In this article, we propose to restrict such projection methods to purely visualization purposes and instead use geometric summary statistics, methods from spatial point statistics, and topological data analysis to characterize the configurations of random programs that do not rely on embedding approximations. To demonstrate their usefulness, we compare two publicly available models: ChatGPT-4 and TinyLlama, on a simple problem related to image processing. 

Application areas include understanding how questions should be asked to obtain useful programs; measuring how consistently a given large language model answers; and comparing the different large language models as a programming assistant. Finally, we speculate that our approach may in the future give new insights into the structure of programming languages.
\end{abstract}

\section{Introduction}
In this article, we consider all the grammatically correct programs of a given programming language as a metric space $(M',d)$, where $M'$ is the discrete set of all possible programs and $d: M'\rightarrow M'\rightarrow \Re$ is a distance function between them. The set $M'$ is countable infinite and embodies all the programs, which can be written, while the distance function $d$ describes some notion of similarity between programs. Given a distance function, we can now construct a fully connected graph $G=(M', E)$ with programs being vertices $m\in M'$ and edges $e\in E$ between programs that have the distance as an attribute. 

For practical reasons, we will consider subsets $M\subset M'$, such that $m\in M$ is a small program related to a limited problem domain, and we consider distances between such programs, $d_{ij} = d(m_i,m_j)$, $m_i,m_j\in M$. With the distance matrix $D = \{d_{ij}\}$, we can apply a suite of standard tools for describing the geometry, topology, and statistics of $M$ as induced by the choice of $d$. For example, if $M$ is a random subset of $M'$, then we may calculate its geometric medians, absolute distance \cite{kuhn73}, its density and tendency to cluster \cite{Ripley1976}, and we can apply topological data analysis tools such as persistent homology diagrams \cite{ChazalFMichelB2021,bobrowski23}. Sometimes it is useful to approximate point sets in low-dimensional Euclidean spaces, e.g., using multidimensional scaling \cite{Mead1992}, however, we find that results obtained using such embeddings can make comparison between sample sets difficult, and therefore, we are in general vary on conclusion derived from embeddings, and thus, we will only use them for visualization.

The analysis of $M$ relies on the choice of distance function $d$. The applicability of the suite of tools, we present here, is independent of the choice of $d$, but the ensuing analysis is not. Commonly used distance functions are based on string matching such as Hamming and edit distances \cite{navarro01}, and which attributes distance to both the spelling and the meaning of programs. In this work, we favor tree-edit distances \cite{pawlik.augsten15,pawlik.augsten16} of parsed grammar trees, since the parse tree is a concise description of the syntax of a program, highlighting programming constructs rather than the spelling of keywords, etc, and even allows for advanced program transformation such as refactoring, etc. In this article, we will give examples in the Python language, as this is one of the dominating languages at the moment and a popular choice for small programs written by non-computer science with the aid of large language models \cite{chatgpt23,llama23}. Python has the further advantage that a typical Python interpreter is equipped with a parsed library, and we use the \texttt{ast}-module provided by the Python foundation \cite{PythonAST2024}.

We use the Python built-in \texttt{ast} module to generate abstract syntax trees. As an example, consider the Python program
\begin{lstlisting}
a = 'Hello World'
print(a)
\end{lstlisting}
and its syntax tree generated by \texttt{ast} shown in \Cref{fig:pythonSyntaxTree}.
\begin{figure*}
\begin{lstlisting}[frame=single]
Module(
  body=[
    Assign(
      targets=[
        Name(id='a',  ctx=Store())],
      value=Constant(value='Hello  World')),
    Expr(
      value=Call(
        func=Name(id='print',  ctx=Load()),
        args=[
          Name(id='a',  ctx=Load())],
        keywords=[]))],
  type_ignores=[])
\end{lstlisting}
\caption{The abstract syntax tree for the program "\lstinline{a = 'Hello World'; print(a)}" as generated by the builtin \texttt{ast} module.}
\label{fig:pythonSyntaxTree}
\end{figure*}
This tree is a one-to-one correspondence to the original program except for some comments and other elements, which do not influence the execution of the program, and can be converted back into a program, which semantically is identical to the original program. In this article, we use tree edit distances between such syntax trees. The tree edit distance between two trees $T_1$ and $T_2$ is the minimal number of operations needed to transform $T_1$ into $T_2$ given a small set of operations, such as relabeling, deleting, and insertion of nodes, and where each operation is associated with a cost. For example, a grossly simplified tree diagram of \Cref{fig:pythonSyntaxTree} is shown in \Cref{fig:pythonSyntaxDiagram}.
\begin{figure}
\begin{center}
\begin{comment}
\begin{forest}
  for tree={
    align=center,
    parent anchor=south,
    child anchor=north,
    font=\sffamily,
    edge={thick},
    l sep+=10pt,
    s sep+=10pt,
  }
  [Module
    [body
      [Assign
        [targets
          [Name [{id='a'}] [{ctx=Store()}]]
        ]
        [value
          [Constant [{value='Hello World'}]]
        ]
      ]
      [Expr
        [value
          [Call
            [func
              [Name [{id='print'}] [{ctx=Load()}]]
            ]
            [args
              [Name [{id='a'}] [{ctx=Load()}]]
            ]
            [keywords []]
          ]
        ]
      ]
    ]
    [type\_ignores []]
  ]
\end{forest}
\end{comment}
\begin{forest}
  for tree={
    align=center,
    parent anchor=south,
    child anchor=north,
    font=\sffamily,
    edge={thick},
    l sep+=10pt,
    s sep+=10pt,
  }
  [body
    [Assign [{id='a'}] [{value='Hello World'}]]
    [Call [{id='print'}] [{id='a'}]]
  ]
\end{forest}
\end{center}
\caption{A much simplified diagram of \Cref{fig:pythonSyntaxTree} for illustration purposes.}
\label{fig:pythonSyntaxDiagram}
\end{figure}
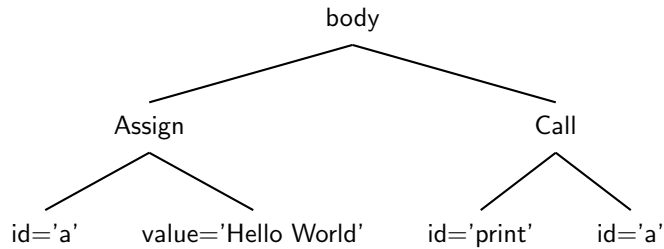
The distance between this tree and another, representing the Python program
\begin{lstlisting}
a = 'Goodbye World'
print(a)
\end{lstlisting}
would be 1, since the leaf representing "value='Hello World'" must be replaced with "value='Goodbye World'". To compute distances between trees, we use \cite{pawlik.augsten15,pawlik.augsten16}\jon{Is this the apted package? Can't find Ken's mail about the Paassen et al. papers}, which has the computational complexity of $\mathcal{O}(n^2)$, where $n= \max(|T_1|,|T_2|)$ and $|T_i|$ is the number of nodes in $T_i$.

Our setup opens up for performing invariant transform of code before generating trees, for example, recursive functions and \lstinline{for} loops may be automatically unfolded or rewritten as a common \lstinline{while} loop, and in this way, design distance functions where different types of loops are considered close to each other. Similarly standard compiler optimization techniques could be a great source of rewriting code that emphasizes semantic identities and given specific target languages and computing architectures for improved speed of execution. Nevertheless, in this paper, we limit ourselves to demonstrating how a suite of geometry, topology, and statistical tools for analyzing point sets can be used to study stochastic program sources and programming languages.

\section{Characterization of point sets from pairwise distances}
In the following, we will summarize the set of tools we use to analyze our metric space $(M,d)$, programs $m_i\in M$, pairwise distances $d_{ij}=d(m_i,m_j)$, and the distance matrix $D=\{d_{ij}\}$. Since $d$ is a distance function, then $D$ is square, symmetric, and has a zero diagonal.

\subsection{Geometric medians and absolute distances}
The geometric median of a set of points in an Euclidean space is the point that minimizes the sum of distances to all points in the set. In contrast to the arithmetic mean, the geometric median is very robust with a breakdown point of 0.5, and on curved spaces, the arithmetic mean is always an element of this space. Thus, it is a good choice for describing the central tendency of program samples. The geometric median $\text{med}$ is defined as:
\begin{equation}
\text{med} = \underset{m_i \in M}{\text{argmin}} \sum_{i\neq j}d_{ij}.
\end{equation}
Finding the geometric median is a non-trivial task, especially as the number of dimensions increases. While there is no simple algebraic solution akin to computing the mean, given the distance matrix $D$, finding the minimum is linear in the number of programs. The median absolute deviation around the geometric median is defined as,
\begin{equation}
\text{mad} = \underset{m \in M}{\text{median}} (m-\text{med}),
\end{equation}
and similarly has a breakdown point of 0.5 and is fast to calculate.

\subsection{Estimating clustering tendencies with Ripley's K-function}
With spatial descriptive statistics, we can describe higher-order features of point sets solely by considering their distances. An example of a descriptor is Ripley's K-function \cite{Ripley1976}, which for a given $n$ points from a random point process, the typical sample-based estimator is defined as,
\begin{equation}
    \hat{K}(r) = \frac{1}{\lambda} \sum_{i\neq j} \frac{I(d_{ij}<r)}{n}.
\end{equation}
Here, $I$ is the indicator function such that $I(x) = 1 \text{ if $x$ is true and otherwise } 0$, and $\lambda$ is the density function. In this article, we will only consider homogeneous densities and set $\lambda=1$ when comparing programs from the same space.

In Euclidean spaces, a homogeneous Poisson point process, henceforth a Poisson process, is often used as a reference process. A Poisson process is a uniformly distributed set of $n$ points, where $n$ is a random Poisson distributed number. The K-function for Poisson processes in $k$ dimensions is known to converge to the volume of the $k$-dimensional ball. Thus, for a Poisson process in 2-dimensions, $K(r) = \pi r^2$, and hence, for an unknown process, in intervals in $r$ where its K-function is observed to be below or above $\pi r^2$, the points are said to be attracting or repelling each other respectively. As an example, consider a garden with apple trees and identify the position where each apple falls in autumn with a point. Since the apples have a finite size, there will be a minimum distance for which these points can lie, i.e., the K-function will show that they repel at small scales. On the other hand, the apples will tend to cluster near the stem of the trees, or equivalently, the apples attract at distances comparable to the crowns of the trees. 

For more complicated spaces, such as programs with a tree-edit distance metric, no simple point process is known, and we do not have the same reference curve. Nevertheless, points estimated from the same or different point process may be compared directly as long as the embedding space is constant. Thus we say that for intervals in $r$, where the K-function increases fast compared to other comparable curves, we say that the programs tend to cluster at the scale of $r$.

\subsection{Topological data analysis on random programs}
Topological data analysis (TDA) is a recently emerged field combining topology, geometry, and statistics on point sets \cite{edelsbrunner02, ChazalFMichelB2021}. TDA considers a nested set of simplicial complexes.

The convex hull of $k+1$ affinely independent points $m_i\in M$ are the vertices of a $k$-dimensional simplex 
\begin{equation}
    \sigma=[m_0,m_1,\ldots,m_k].
\end{equation}
Any subset of the vertices of $\sigma$ is a lower dimensional simplex and is called its face. For example, the convex hull of 3 affinely independent points $\{m_0,m_1,m_2\}$ is the 2-dimensional triangle $[m_0,m_1,m_2]$ and its faces are its 1-dimensional lines $[m_0,m_1]$, $[m_1,m_2]$, $[m_2,m_0]$, who in turn have 0-dimensional faces, e.g., the faces of $[m_0,m_1]$ is $[m_0]$ and $[m_1]$.

A simplicial complex $S$ is a set of simplices such that
\begin{enumerate}
    \item if $\sigma_i\in S$, then its faces are also in $S$, and
    \item the intersection between two simplices is empty or a common face of the two. 
\end{enumerate}
For example, the set of point and line simplicies $L=\{[m_0],[m_1],[m_2],[m_0,m_1], [m_1,m_2], [m_2,m_0]\}$ is a simplicial complex. 

The $k$-dimensional homology of a simplicial complex $H_k(S)$ is a vector space whose basis is related to the $k$-dimensional cycles of the complex, that is, $H_0$ represents the connected components of $S$, $H_1$ the 1-dimensional holes in $S$, $H_2$ the number of 2-dimensional hollows in $S$, etc. Here we follow \cite{ChazalFMichelB2021}, and define the $k$-dimensional homology for simplicial complexes using $k$-chains with coefficients in the group $\mathbb{Z}_2$. For the $k$-dimensional simplices of a simplicial complex $S$, $\{\sigma_0,\sigma_1,\ldots,\sigma_{n-1}\}$, its $k$-chain is formally written as $c_k = \sum_{i=0}^{n-1} z_i\sigma_i$. Only identical simplices can be added using the rules of $\mathbb{Z}_2$, where $\mathbb{Z}_2=\{0,1\}$ with cyclic addition implying that besides regular integer addition, we have that $1+1=0$, $0-1=1$, and the product is defined as usual for integers. As a consequence given a scalar $z'\in\mathbb{Z}_2$ scalar multiplication is given as $z'c = \sum_{i=0}^{n-1} z'z_i\sigma_i$, and given another chain, $c'=\sum_{i=0}^{n-1} z_i'\sigma_i$, addition is given by $c+c'=\sum_{i=0}^{n-1} (z_i+z_i')\sigma_i$. By this, we see that the set of all $k$-dimensional chains $C_k(S)=\{c_i\}$ is a vector space with coefficients in $\mathbb{Z}_2$. The $k$-dimensional boundary of a $k$-dimensional simplex is the $k-1$-dimensional chain, $\partial_k [m_0,\ldots,m_k] = \sum_{i=0}^k(-1)^i[m_0,\ldots,m_{i-1},m_{i+1}\ldots,m_k]$, and it is extended to chains as $\partial_k c = \sum_{i=0}^{n-1} z_i\partial_k \sigma_i$. Using the boundary operator, we can define the kernel, $\texttt{kern}_k(S) =\{c\in C_k(S): \partial_k c=0\}$, and the image $\texttt{im}_k(S) = \{c \in C_k(S): \exists c' \in C_{k+1}(S), \partial_{k+1} c' = c\}$. Thus we can define the $k$-dimensional homology as the quotion space, 
\begin{equation}
H_k(S) = \frac{\texttt{kern}_k(S)}{\texttt{im}_k(S)}.
\end{equation}

The dimensionality of $H_k$ is called the Betti number, $\beta_k=\texttt{dim}(H_k)$, and is directly related to the number of $k$-dimensional cycles of $S$. That is, $\beta_0$ is the number of connected components, and $\beta_n, n>0$ is the number of n-dimensional cycles. Naturally, the best number for dimensions higher than the largest dimension of the simplices in $S$ is 0.  Thus, continuing the example from the above, $\beta_0(S_0) = 1$, $\beta_0(S_1) = 1,\beta_1(S_1) = 1$, and $\beta_0(S_2) = 1,\beta_1(S_2) = 0,\beta_2(S_2) = 1$. 

A sequence of simplical $[S_0,S_1,\ldots]$ is said to be nested if $S_i\subseteq S_j, i<j$. For example, $S_0=\{[m_0],[m_1],[m_2]\}$, $S_1= L$, and $S_2= L\cup [m_0,m_1,m_2]$ is a nested sequence. Such nestings are called filtration, and a popular filtration for metric point sets is the Vietoris-Rips filtration in which for a given scale $r$, a simplex $\sigma=[m_0,m_1,\ldots,m_k]$ is in $S_r$ if for all pairs in the simplex, $d(m_i,m_j)\leq r$. Topological features of the nested sequences may persists over several complices $S_i$, and tracking $k$-cycles along scale gives rise to the notion of birth and death of a $k$-cycle. A connected component or $k$-cycles, $k>0$ are said to be born and die at the index of the complex $S_i$ where they first and last are observed. Connected components are treated differently than other cycles. They are initially points and are born at $0$, when two components are connected, then that which appeared last dies, and at $\infty$ all points will be connected into a single component. Plotting the distribution of (birth,death) for a filtration is known as the persistence diagram. Further, the number of $k$-cycles in $S_i$ is equal to $\beta_k(S_i)$, and the stepwise constant function of Betti numbers by scale is known as Betti curves.

\jon{Add a short description of \cite{bobrowski23}.}

\subsection{Projection methods}
Multidimensional scaling \cite{Mead1992} and the eigenvector space of graph laplacians \cite{godsil.royle01} can be used to visualize high dimensional distance matrices in lower dimensions. Analyses such as detecting clusters etc.\ in such projections should take into account the resulting uncertainty wrt.\ the discarded information. The weighted graph laplacian, $L$, is for our case found to be $L=|M|I-D$, where $I$ is an $|M|x|M|$ identity matrix, and projecting the points in $M$ is a popular data reduction method. Here we will focus on multidimensional scaling, which seeks a set of points in $p_i,p_j\in\Re^n$, whose pairwise Euclidean distances are as close as possible to a given set of distances $d_{ij}$. Thus setting, e.g., $n=2$ gives a point cloud of similar distances but where the cloud's translation and rotational values are irrelevant.

\section{Random programs from Large Language Models}
As a source of random programs, we probed two different large language models: ChatGPT-4 \cite{openai_chatgpt4} and TinyLlama \cite{zhang.zeng.wang.lu24}. ChatGPT-4 is a private model pay-per-query model made available by the non-profit OpenAI, Inc. company and its for-profit subsidiary OpenAI Global, LLC. What is publicly known about the model is that it is a Generative pre-trained transformers (GPT) large language model and it is estimated to have 1.76 trillion parameters \cite{wikipedia_gpt4}. TinyLlama builds on the architecture and tokenizer of \cite{touvron.ea23,} and has 1.1 billion parameters.

Both models use the concept of tokens as the smallest language unit that the model is training on and can make predictions on. The implementations vary between models, and OpenAI does not disclose how they tokenize sentences but do provide a tokenizer tool \cite{openai_tokens}, which for chatGPT-3.5 and -4 demonstrates that the sentence "Hello, world!", is tokenized as ["Hello", ",", "world", "!"]. TinyLlama uses bytepair encoding \cite{sennrich.haddow15,kudo.richardson18}

\begin{figure}
    \centering
    \begin{subfigure}{0.45\linewidth}
        \includegraphics[width=\textwidth]{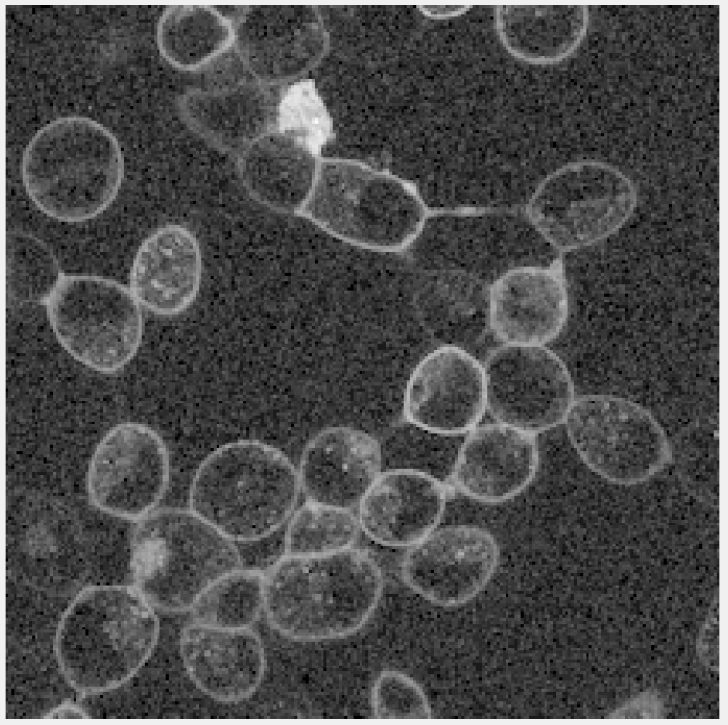}
        \caption{}
        \label{fig:image}
    \end{subfigure}
    \hfill
    \begin{subfigure}{0.45\linewidth}
        \includegraphics[width=\textwidth]{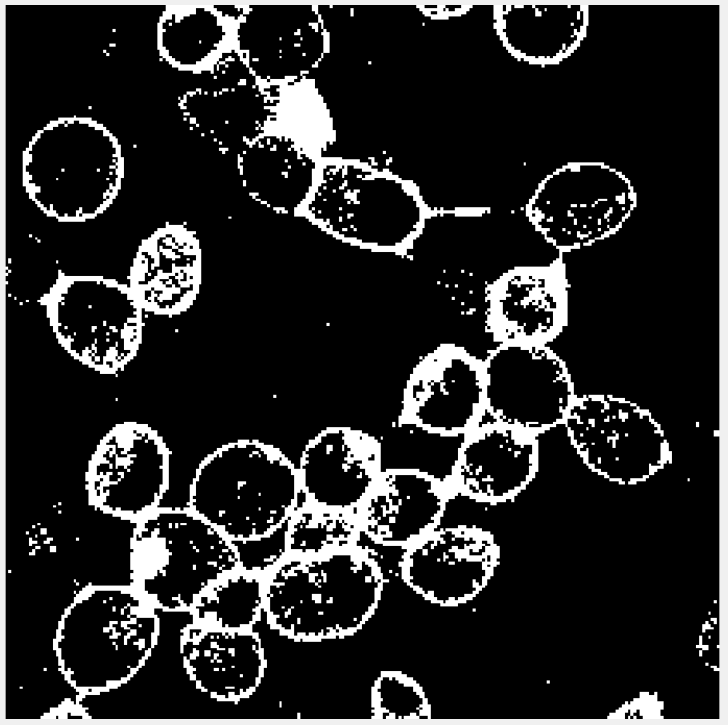}
        \caption{}
        \label{fig:threstholded}
    \end{subfigure}
    \caption{\csubref{fig:image} A microscope image of cells and \csubref{fig:threstholded} its thresholded version. The circular features are cells, and thresholding mainly classifies the pixels into cell walls or not. The image courtesy of Karen Martinez and Gabriella von Scheel von Rosing, University of Copenhagen.}
    \label{fig:threshold}
\end{figure}

To compare models we chose a simple application area very familiar to us: Segmenting 2-dimensional images with thresholding. Thresholding is a binary pixel-classification task, where for a given image $J$, coordinate $\vec x$, and threshold value $t$, each pixel is classified as $I(\vec x)>t$. An example is illustrated in \Cref{fig:threshold}.
A popular choice for selecting the threshold value is Otsu's method \cite{otsu79}. Further pre- and post-processing is often needed such as noise reduction before segmentation and object identification after. Modern segmentation techniques typically rely on deep-learning models including the values of the neighboring pixels, however, we find that for studying the distribution of programs generated by large language models, queries generating small programs are more enlightening.

In this article, we do not probe deeply into the relation between human language queries and program responses, but from our personal experience teaching students to program, we identified some key features of a query: 
\begin{itemize}
    \item Code encapsulation,
    \item Encapsulation interface,
    \item Solution specification, and
    \item Pretended behavior of the model.
\end{itemize}
Thus our queries focus on whether to ask for a function or a program, the precision with which the input and output are specified, and to which degree a specific thresholding method is requested or not.

Both models answer queries with a mix of English text and program snippets. For homogeneous treatment, we concatenated program snippets into a single program. Since some snippets included installation instructions, this procedure did not always result in a syntactically correct Python program. These responses were ignored.

\section{Experiments}
The key features of a query lead us to formulate the 7 questions shown in \Cref{tab:questions}.
\begin{table}
    \centering
\begin{tabular}{c|p{0.82\linewidth}}
     Id &  Question\\\hline
     0 & Write a \underline{function} which thresholds an image.\\
     1 & Write a \underline{function} which segments an image using thresholding.\\
     2 & Write a \underline{program} which thresholds an image.\\
     3 & Write a python \underline{function} that segments an image using \underline{Otsu's} threshold from \underline{opencv}. The \underline{function} must check that the input is a \underline{gray value 2d np.arrays}, and it is not to perform \underline{noise reduction}.\\
     4 & Write a \underline{function} which takes an \underline{image as input} and \underline{returns a segmentation} using \underline{Otsu} thresholding.\\
     5 & Write a \underline{function} which takes an \underline{image as input} and \underline{returns a segmentation} using thresholding.\\
     6 & Act as an \underline{experienced python programmer}. Write a \underline{function} which takes an \underline{image as input} and \underline{returns a segmentation} using \underline{Otsu} thresholding.
\end{tabular}
    \caption{The language models were asked 7 different questions for producing a program or a function, which segments an image. Underlining highlights the key query features beyond "thresholding an image".}
    \label{tab:questions}
\end{table}

For the experiments, we used a MacBook Pro with an Apple M1 chip, 16 GB of memory, and running MacOS v.\ 14.5. Each model was asked the same question independently 100 times, which for each session, resulted in $701*700/2=254.350$ queries. 

We asked ChatGPT-4 using openAI's team interface at its default temperature. The code parts of each response were extracted and concatenated in the order they were given. The Ast python parser was not able to parse 65 out of 700 programs due to syntax errors. The main computational model was the calculations of the distances on our MacBook. The distance matrix was computed in about 7h. 

TinyLlama was installed on our MacBook Pro The questions were repeated 100 times, and 3 different temperatures were probed: 0.5, 0.7, and 0.9. The response time for each question was in the order of 10 seconds. The code parts of each response were extracted and concatenated in the order they were given. The Ast python parser was not able to parse all programs due to syntax errors, and the (temperature, \#errors) were: (0.5,3), (0.7,12), and (0.9,23). The erroneous programs were discarded. The major bottleneck for analyzing the resulting programs was calculating the distance matrix. With 3 parallel processes calculating the distance matrices in Python, we were able to compute in the order of 12.000 distances per hour and the three distance matrices were calculated in approximately 70h. 

\section{Results}
The responses were analyzed per model and for each model we analyzed the questions separately. The resulting programs were typically in the order of 10-20 lines long, and for a randomly select subset, their quality were assessed by the authors reading the programs. Typical variations were whether the input to the snippets were filenames from which to read the image or the actual images, whether the images were assumed to be gray-value or color, whether smoothing should be performed prior to thresholding, and which python package to be used for thresholding. No semantical errors were discovered.

Table \Cref{tab:summaryStat} shows the summary statistics for all 4 models. 
\begin{table*}
    \centering
\begin{tabular}{cc|cccccccc}
\multirow{2}{4em}{Model} & \multirow{2}{4em}{Statistics} & \multicolumn{8}{l}{Group} \\
 & & 0-6 &  0 &  1 &  2 &  3 &  4 &  5 &  6 \\\hline
ChatGPT-4 & Mid Prg Idx & 442 & 1 & 141 & 221 & 344 & 451 & 545 & 621 \\
&Avg Dispersion & 65.0 & \textbf{42.5} & 48.1 & 55.0 & 49.1 & 60.0 & 43.4 & 54.5 \\
&Median Dispersion & 57.0 & 34.0 & 41.5 & 45.0 & 47.0 & 45.0 & \textbf{31.5} & 43.0 \\
&Average Stress & 195.5 & 83.4 & \textbf{79.5} & 102.7 & 100.2 & 122.1 & 80.7 & 97.5 \\\hline
TinyLlama $t=0.5$ & Mid program idx & 472 & 23 & 88 & 22 & 47 & 25 & 54 & 32 \\
&avg dispersion & 148.3 & 110.7 & 220.6 & \textbf{68.5} & 167.3 & 120.9 & 131.9 & 151.6 \\
&median dispersion & 121.0 & 103.0 & 161.5 & \textbf{38.5} & 136.0 & 120.0 & 116.5 & 138.0 \\
&average stress & 1495.8 & 661.1 & 2034.4 & \textbf{248.9} & 1966.0 & 896.1 & 983.9 & 1738.3\\\hline
TinyLlama $t=0.7$ & Mid program idx & 256 & 3 & 4 & 45 & 44 & 87 & 1 & 22 \\
&avg dispersion & 167.4 & 138.5 & 215.9 & \textbf{88.9} & 184.9 & 162.0 & 155.3 & 166.4 \\
&median dispersion & 138.0 & 122.0 & 169.0 & \textbf{59.0} & 163.0 & 134.5 & 126.0 & 140.0 \\
&average stress & 1927.7 & 1187.7 & 2637.9 & \textbf{510.4} & 2241.7 & 1528.0 & 1249.5 & 1953.3\\\hline
TinyLlama $t=0.9$ & Mid program idx & 111 & 80 & 15 & 70 & 59 & 41 & 7 & 26 \\
&avg dispersion & 207.7 & 196.1 & 256.0 & \textbf{107.3} & 204.2 & 215.1 & 229.6 & 225.3 \\
&median dispersion & 155.0 & 160.0 & 199.5 & \textbf{65.0} & 159.0 & 175.0 & 155.0 & 186.0 \\
&average stress & 2910.4 & 2312.3 & 3890.5 & \textbf{554.4} & 2391.6 & 3005.7 & 2341.4 & 4069.2
\end{tabular}
    \caption{ChatGPT-4 responses have a smaller dispersion than TinyLlama, and TinyLlama's dispersion increases with temperature. The smallest value per row is highlighted in bold. The Mid program index (Mid Prg Idx) refers to one of the unique indices in the set of 700 queries per session. The same indices are used in the projections shown in \Crefrange{fig:chatGPT4}{fig:tinyllama09}.}
    \label{tab:summaryStat}
\end{table*}

\begin{figure*}
\centering
\begin{tabular}{c|cccc}
 & ChatGPT-4 & TinyLlama $t=0.5$ & TinyLlama $t=0.7$ & TinyLlama $t=0.9$\\\hline
\rotatebox{90}{Embeddings}
& \includegraphics[width=0.2\linewidth]{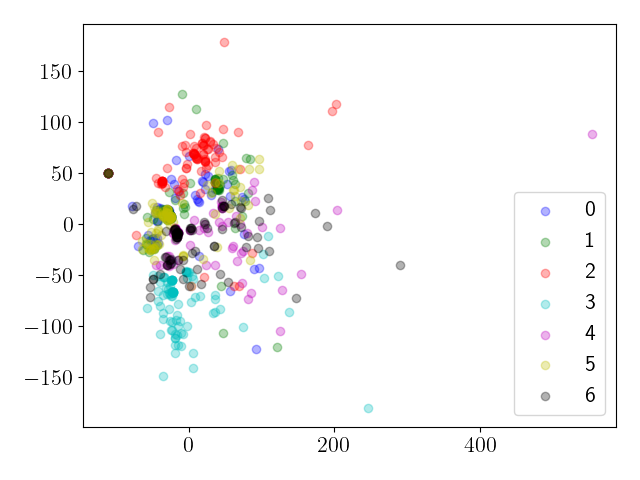}
& \includegraphics[width=0.2\linewidth]{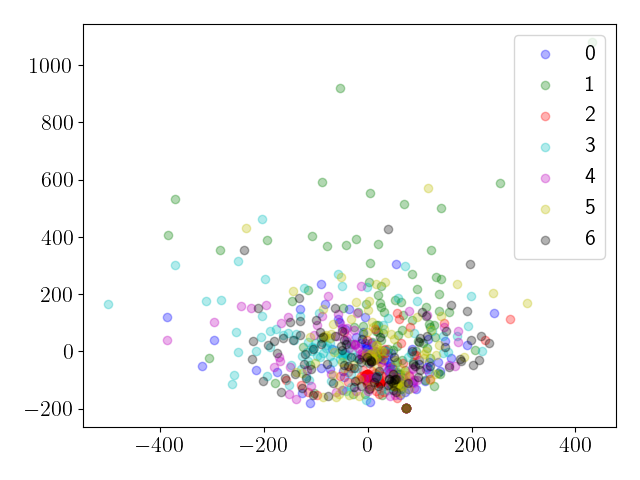}
& \includegraphics[width=0.2\linewidth]{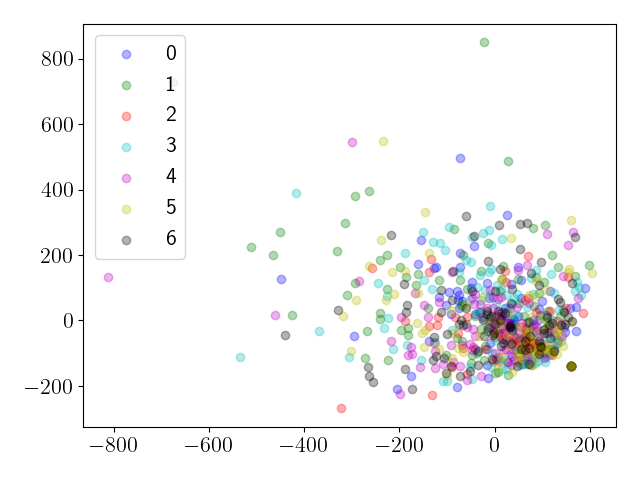}
& \includegraphics[width=0.2\linewidth]{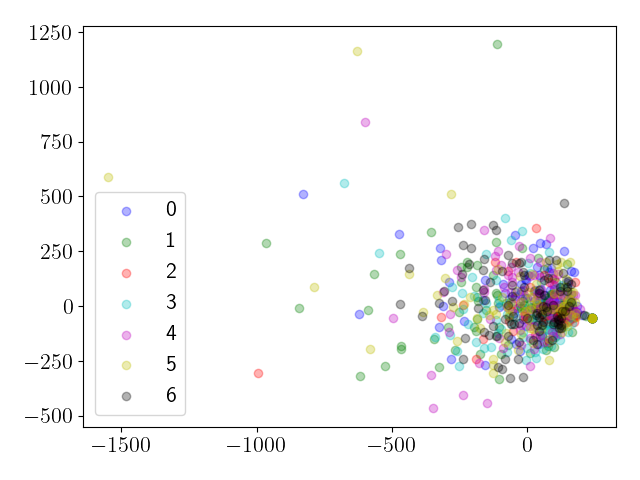}\\
\rotatebox{90}{K-function}
& \includegraphics[width=0.2\linewidth]{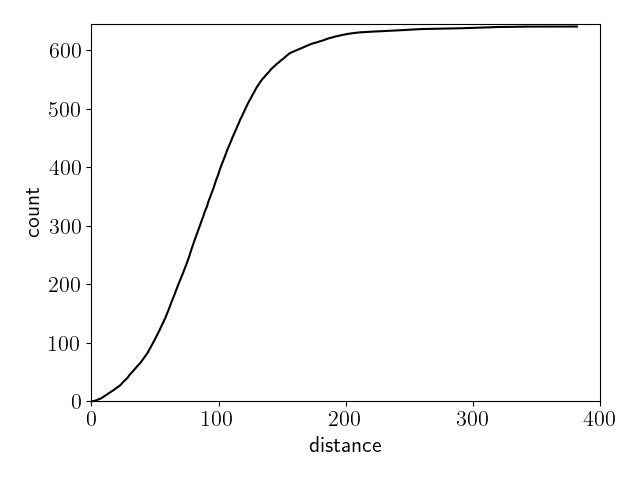}
& \includegraphics[width=0.2\linewidth]{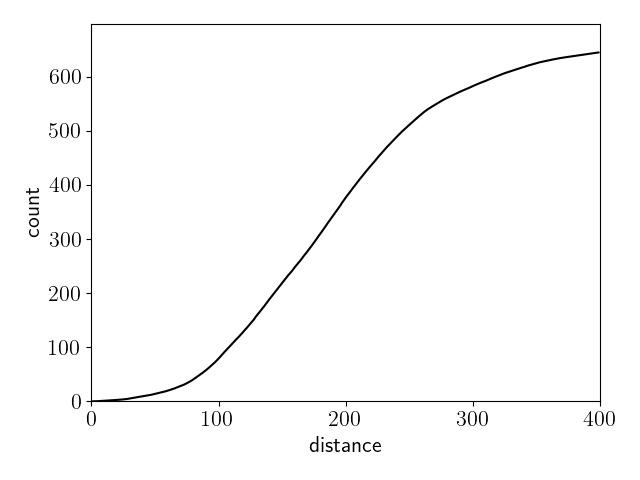}
& \includegraphics[width=0.2\linewidth]{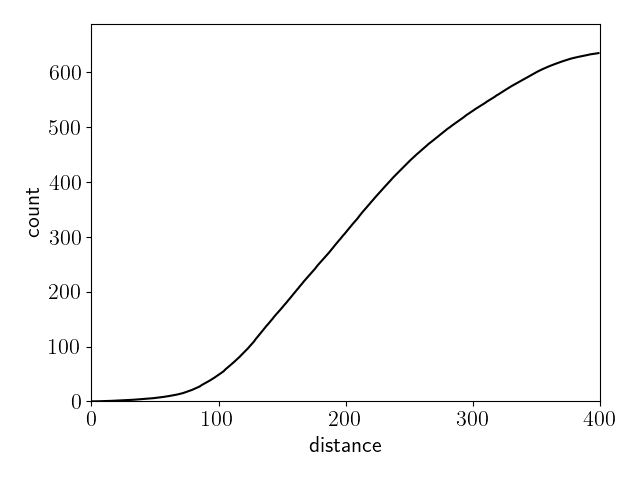}
& \includegraphics[width=0.2\linewidth]{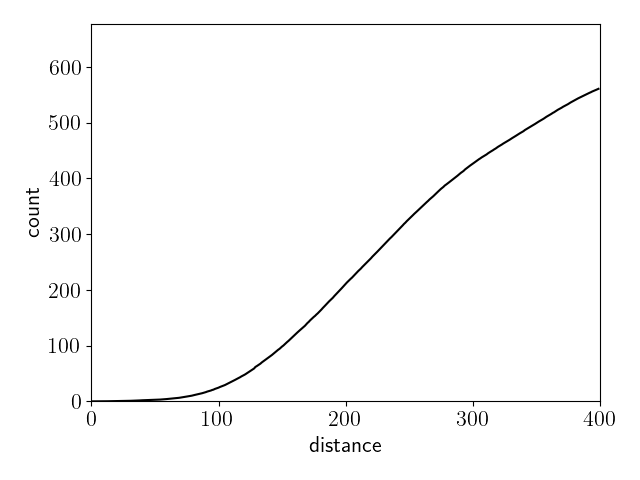}\\
\rotatebox{90}{Persistence dgm}
& \includegraphics[width=0.2\linewidth]{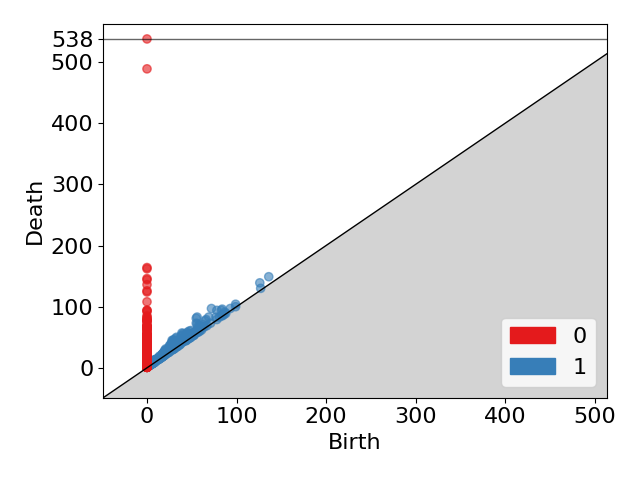}
& \includegraphics[width=0.2\linewidth]{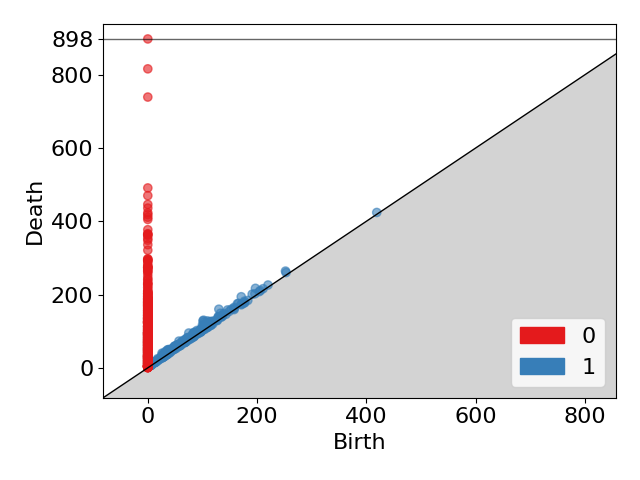}
& \includegraphics[width=0.2\linewidth]{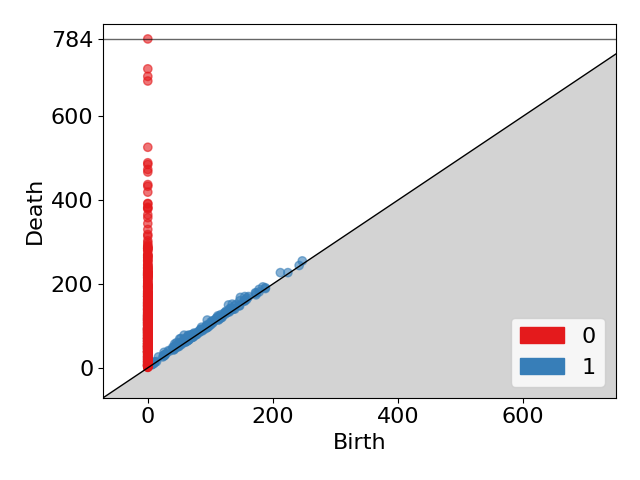}
& \includegraphics[width=0.2\linewidth]{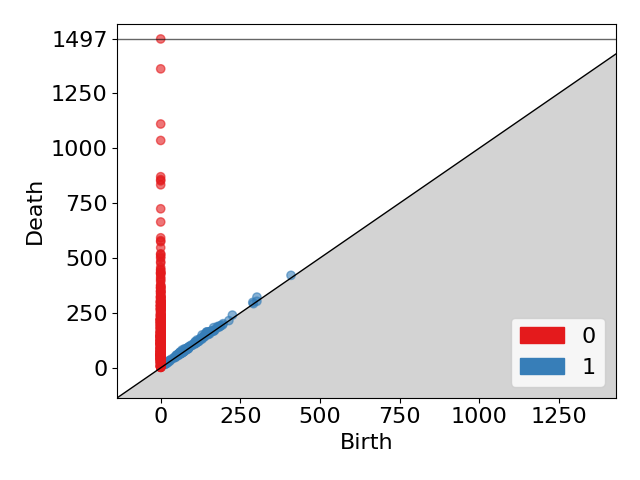}\\
\rotatebox{90}{Log diagrams}
& \includegraphics[width=0.2\linewidth]{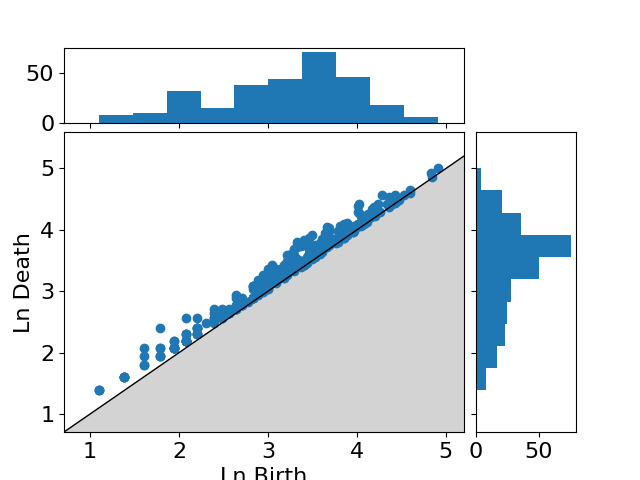}
& \includegraphics[width=0.2\linewidth]{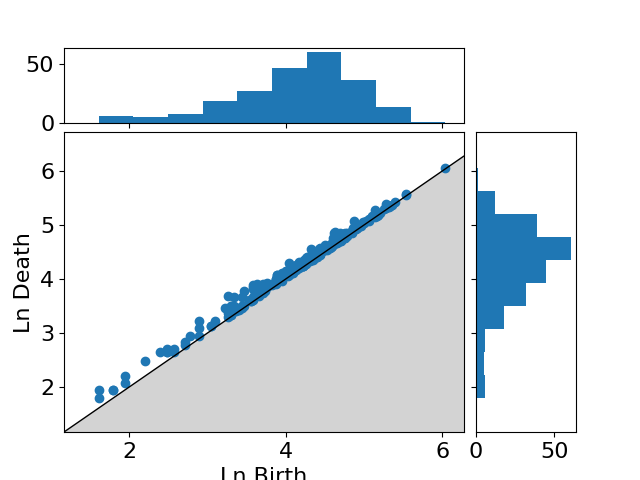}
& \includegraphics[width=0.2\linewidth]{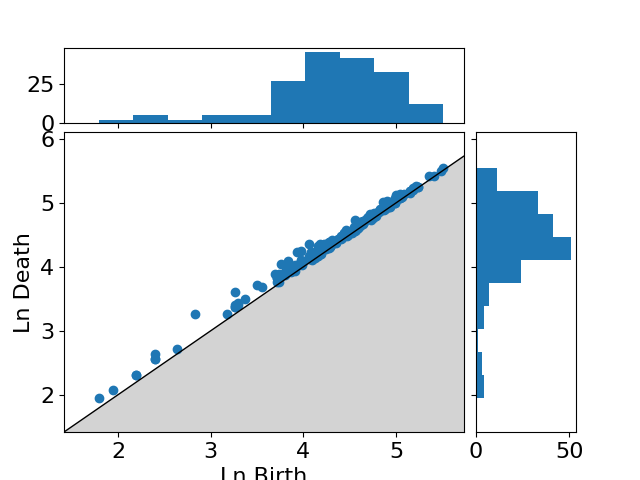}
& \includegraphics[width=0.2\linewidth]{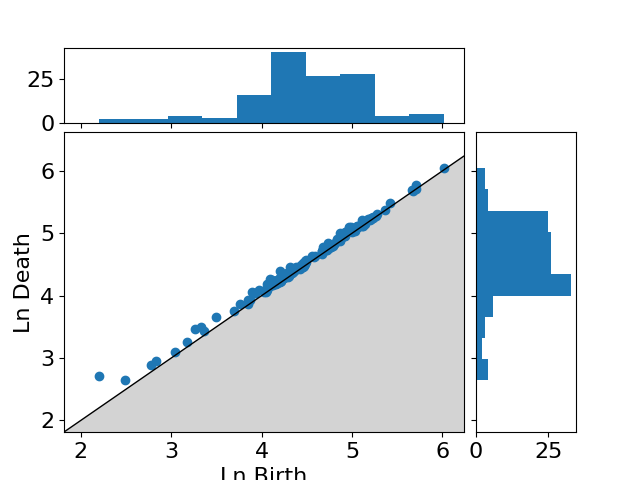}
\end{tabular}
\caption{A summary of \Crefrange{fig:chatGPT4}{fig:tinyllama09} in the Supplementary Materials section shows the results of analyzing the responses from all the queries in a session. In the embeddings, the colors refer to the question numbers in \Cref{tab:questions}. ChatGPT-4's responses seem more clustered than TinyLlama's responses when the programs are projected into 2 dimensions using multidimensional scaling. Similarly for the K-functions, ChatGPT-4 responses are more clustered than TinyLlama, and TinyLlama's dispersion increases with temperature. In the persistence diagram, red and blue dots illustrate the birth and death coordinates of the connected components and 1-cycles respectively. For ChatGPT-4 their scales are smaller than TinyLlama's, whose dispersion increases with temperature. The log persistence diagrams show additionally the marginal histograms, and there is perhaps a tendency for the distribution of ChatGPT-4's 1-cycles to be more widely spread than TinyLlama's.}
\label{fig:summary0-6}
\end{figure*}

The projections, Ripley's K-functions, the persistence diagrams diagram, and their log-diagrams for the 1-cycles for each session divided into all responses and per query responses are shown in the supplementary materials section in \Crefrange{fig:chatGPT4}{fig:tinyllama09}. The figures illustrating the behavior of all the questions per model is repeated in \Cref{fig:summary0-6}.

\section{Discussion}
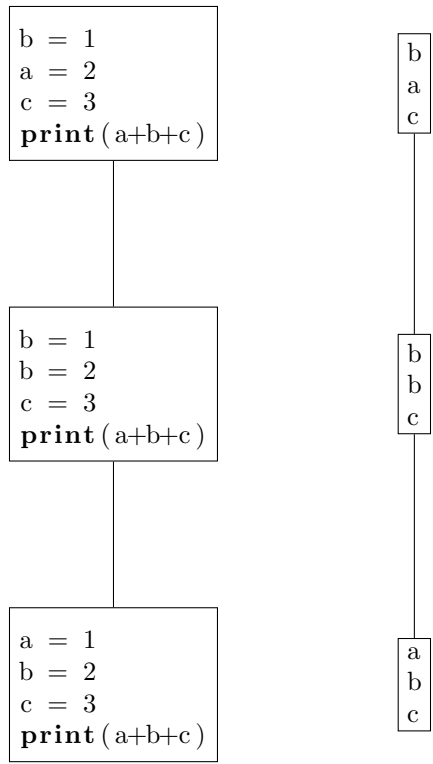
\begin{figure}
    \centering
\begin{tikzpicture}
\node[shape=rectangle,draw=black] (ABC) at (0,0) {
\begin{lstlisting}
a = 1
b = 2
c = 3
print(a+b+c)
\end{lstlisting}
    };
\node[shape=rectangle,draw=black,align=left] (abc) at (4,0) {a\\b\\c};
    \node[shape=rectangle,draw=black] (BBC) at (0,4) {
\begin{lstlisting}
b = 1
b = 2
c = 3
print(a+b+c)
\end{lstlisting}
    };
\node[shape=rectangle,draw=black,align=left] (bbc) at (4,4) {b\\b\\c};
    \node[shape=rectangle,draw=black] (BAC) at (0,8) {
\begin{lstlisting}
b = 1
a = 2
c = 3
print(a+b+c)
\end{lstlisting}
    };
\node[shape=rectangle,draw=black,align=left] (bac) at (4,8) {b\\a\\c};

    \path [-] (ABC) edge node[left] {} (BBC);
    \path [-] (BBC) edge node[left] {} (BAC);
    \path [-] (abc) edge node[left] {} (bbc);
    \path [-] (bbc) edge node[left] {} (bac);
\end{tikzpicture}
    \caption{Interchanging variable names \lstinline{a} and \lstinline{b} is a 2 tree-edit distance transformation. Left shows a minimal path and right its shorthand.}
    \label{fig:transitionShorthand}
\end{figure}

\begin{figure}
    \centering
\begin{tikzpicture}
    \node[shape=rectangle,draw=black,align=left] (abc) at (0,1) {a\\b\\c};
    \node[shape=rectangle,draw=black,align=left] (aac) at (1,0) {a\\a\\c};
    \node[shape=rectangle,draw=black,align=left] (bbc) at (1,2) {b\\b\\c};

    \node[shape=rectangle,draw=black,align=left] (bac) at (2,1) {b\\a\\c};
    \node[shape=rectangle,draw=black,align=left] (baa) at (3,0) {b\\a\\a};
    \node[shape=rectangle,draw=black,align=left] (bcc) at (3,2) {b\\c\\c};

    \node[shape=rectangle,draw=black,align=left] (bca) at (4,1) {b\\c\\a};
    \node[shape=rectangle,draw=black,align=left] (bba) at (4,3.5) {b\\b\\a};
    \node[shape=rectangle,draw=black,align=left] (cca) at (5,3.5) {c\\c\\a};

    \node[shape=rectangle,draw=black,align=left] (cba) at (4,6) {c\\b\\a};
    \node[shape=rectangle,draw=black,align=left] (caa) at (3,5) {c\\a\\a};
    \node[shape=rectangle,draw=black,align=left] (cbb) at (3,7) {c\\b\\b};

    \node[shape=rectangle,draw=black,align=left] (cab) at (2,6) {c\\a\\b};
    \node[shape=rectangle,draw=black,align=left] (aab) at (1,5) {a\\a\\b};
    \node[shape=rectangle,draw=black,align=left] (ccb) at (1,7) {c\\c\\b};

    \node[shape=rectangle,draw=black,align=left] (acb) at (0,6) {a\\c\\b};
    \node[shape=rectangle,draw=black,align=left] (abb) at (-1,3.5) {a\\b\\b};
    \node[shape=rectangle,draw=black,align=left] (acc) at (0,3.5) {a\\c\\c};

    \path [-] (abc) edge node[right] {} (abb);
    \path [-] (abc) edge node[right] {} (acc);

    \path [-] (abc) edge node[left] {} (aac);
    \path [-] (abc) edge node[left] {} (bbc);

    \path [-] (bac) edge node[right] {} (aac);
    \path [-] (bac) edge node[right] {} (bbc);

    \path [-] (bac) edge node[left] {} (bcc);
    \path [-] (bac) edge node[left] {} (baa);

    \path [-] (bca) edge node[right] {} (bcc);
    \path [-] (bca) edge node[right] {} (baa);

    \path [-] (bca) edge node[left] {} (cca);
    \path [-] (bca) edge node[left] {} (bba);

    \path [-] (cba) edge node[right] {} (cca);
    \path [-] (cba) edge node[right] {} (bba);

    \path [-] (cba) edge node[left] {} (caa);
    \path [-] (cba) edge node[left] {} (cbb);

    \path [-] (cab) edge node[right] {} (cbb);
    \path [-] (cab) edge node[right] {} (caa);

    \path [-] (cab) edge node[left] {} (ccb);
    \path [-] (cab) edge node[left] {} (aab);

    \path [-] (acb) edge node[right] {} (ccb);
    \path [-] (acb) edge node[right] {} (aab);

    \path [-] (acb) edge node[left] {} (abb);
    \path [-] (acb) edge node[left] {} (acc);
\end{tikzpicture}
    \caption{All permutations of interchanging variable names \lstinline{a}, \lstinline{b}, and \lstinline{c} gives a 1-cycle in tree edit distance for filtrations of radius 1. See \Cref{fig:transitionShorthand} for shorthand notation used.}
    \label{fig:existence1Cycle}
\end{figure}
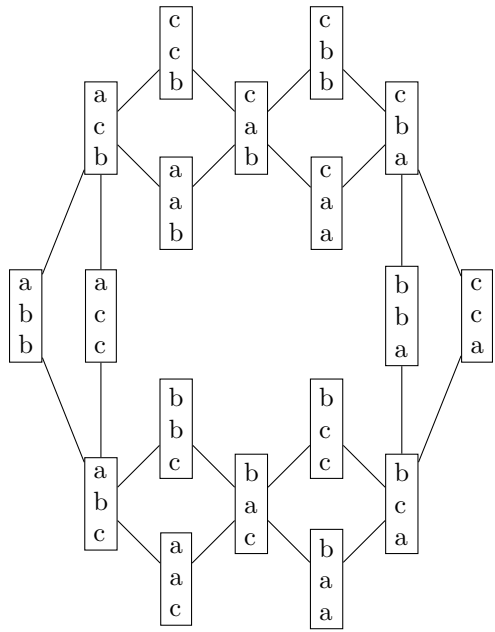

In the following, we will first turn our attention to the full set of responses per model, Group 0-6 in \Cref{tab:summaryStat} and \Cref{fig:summary0-6}. The responses from ChatGPT-4 are much more compactly distributed than the TinyLlama models, and the distribution of responses from TinyLlama increases with temperature. The Average stress of performing multidimensional scaling shows a similar tendency, we are unable to ascertain by these experiments whether the ChatGPT-4 model space is more linear than the TinyLlama, since their spread is so different and may be the reason why the stress is high for TinyLlama. The average and median dispersion differ in the range of 15\%-25\% which indicates the existence of some, non-dominating outliers The K-functions evaluated demonstrate that ChatGPT-4 clusters answers as compared to the other models, which seem to disperse much more. We speculate that this will give improved opportunities for control for focused groups, such as students at various levels of learning since clusters imply that context supplements will likely steer the answers in a useful direction. On the other hand, for TinyLlama, the K-functions show a repulsion tendency near zero scale indicating that all the answers differ, which may be useful for more explorative investigations. This is also indicated by the projection images, where the ChatGPT-4 responses seem to cluster in contrast to all the TinyLlama models. Similarly, the connected components (0-cycles) in the persistence diagrams also indicate that the ChatGPT-4 answers are closer to each other than the TinyLlama answers and that the TinyLlama spread in responses increases with temperature. The 1-cycles show a similar concentration at low scales for ChatGPT-4 but do not seem an increase in spreading for TinyLlama with increased temperature. Further, the log persistence diagrams for 1-cycles are skewed towards larger scales, and perhaps TinyLlama's 1-cycles are more concentrated. No outliers are detected indicating that the 1-cycles are likely noise. This is somewhat surprising since Python language constraints imply that cycles exist. For example consider the program in \Cref{fig:exampleProgram}.
\begin{figure}
    \centering
\begin{lstlisting}
a = 1
b = 2
c = 3
print(a+b+c)
\end{lstlisting}
    \caption{A program for adding 3 values.}
    \label{fig:exampleProgram}
\end{figure}
Interchanging the names \lstinline{a} and \lstinline{b} using the tree-edit distance rules could be as indicated in \Cref{fig:transitionShorthand}.
Consider now all possible permutation of the names \lstinline{a}, \lstinline{b}, and \lstinline{c}, we get possible paths shown in \Cref{fig:existence1Cycle}.
That is, at scale 1, the 18 different syntactically correct programs form a 1-cycle, and this will first close at a later scale. However, it is very likely, that such structures did not appear randomly in our sampling. The existence of 1-cycles in Python leads us to speculate that topological descriptors may be a useful tool to describe languages in general, but not to describe random program sources.

Looking at the per query columns in \Cref{tab:summaryStat} we see that for the TinyLlama models, query 2 has consistently the most compact responses within the TinyLlama model queries, which is quite surprising since this is the only question, which asks for a program rather than a function. Nevertheless, the ChatGPT-4 responses all indicate a more compact distribution per question. However, none of the models have query 3 as the most compact distribution. This is a surprise to us since this is the query with the most detailed specification of the desired response. We speculate that this is a consequence of the queries the models have been trained on since detailed specifications are likely uncommon on fora such as Stack Overflow (\url{https://stackoverflow.co/}). Considering the per query results in \Crefrange{fig:chatGPT4}{fig:tinyllama09} in the Supplementary Material, we observe that for ChatGPT-4, the projections indicate that some responses could show clustering behavior, e.g., the projection for response 0, 1, and 4, however, these are not visible in the corresponding K-functions. None of the projections for the TinyLlama models hints at clustering effects, but the K-functions show great variability in their tendency to repel at low scales, e.g., TinyLlama $t=0.5$ query 1 versus query 2. The K-functions also indicate that the shape of the distribution of responses varies, e.g., response 0 has an initial sharp rise while response 6's rise seems more uniform. In all cases, the individual persistence and log persistence diagrams contain relatively few points, and we hesitate to make substantial conclusions based on them.

Underlying our analyses is the distance function. Here we have used the tree-edit distance of syntactically parsed Python programs. Although the \texttt{ast}-module is very fast and only grows quadratically with the size of the programs being compared, the size of the distance matrix also grows quadratically in the number of programs, we have found that computation time needed is substantial, and we are considering parallel implementations for the application of our work to compare, e.g., student responses. 

Tree-edit distances of syntactically parsed programs are preferred over character distances since it is closer in tune with the meaning of a program, and they better allow for defining distance functions that reflect semantic similarities. For example, the program in \Cref{fig:exampleProgram} uses non-descriptive names, which matters little to the computer but for humans, this does not communicate the intention behind the calculation and is in general not a good programming style. Alternatively, the following program is identical from a syntactical point of view but clearly states the intention behind it,
\begin{lstlisting}
Hours_of_sun_January = 1
Hours_of_sun_February = 2
Hours_of_sun_March = 3
print(Hours_of_sun_January\
    +Hours_of_sun_February\
    +Hours_of_sun_March)
\end{lstlisting}
With parsed syntax trees, it is relatively simple to define a tree-edit distance measure for which these two programs are identical, e.g., by performing automatic refactoring of variable names into an internal and common naming convention. More complicated transformations of programs are possible, e.g., the following recursive loop,
\begin{lstlisting}
def count(n):
  print(n)
  if n > 0:
    count(n-1)
count(2)
\end{lstlisting}
performs the same calculation as,
\begin{lstlisting}
for i in reversed(range(3)):
   print(i)
\end{lstlisting}
In many compilers, such loops are detected and translated to while loops, and thus, distance functions can be defined based on the distances between the translated programs. 

\section{Conclusion}
In this article, we have analyzed a small subset of programs generated by random sources. The programs are all small, and our experience outside what is reported here has led us to believe that the tools used are most useful for small programs, since the set of possible programs grows exponentially with their size. Hence, we view the material presented here as a step in the direction of locally describing programming languages and compact distributions.

Our observations lead us to conclude that summary statistics is a powerful tool to analyze the global distribution of randomly generated programs and that while the projection has visual appeal, the K-function is a better tool to assess the attraction and repulsion effects of. In contrast, topological data analysis did not offer new insights into their distributions. Topological features tend to be very sensitive to noise, but even with recent statistical developments \cite{bobrowski23}, these methods require more developments before we may use them to expand our understanding of random programs. However, as demonstrated in the discussion, the space of Python programs has 1-cycles, and perhaps comparing such topological properties of the space of programs will in the future highlight key differences between programming languages.

\appendix
\section{Supplementary material}
\begin{figure*}
\centering
\begin{tabular}{c|cccc}
Group & Projection & Ripley's K-function & Persistence diagram & Logarithmic Persistence \\\hline
0-6
&\includegraphics[width=0.2\textwidth]{fig/embedding0-6_chatgpt4_0.png}
%&\includegraphics[width=0.2\textwidth]{fig/stress0-6_chatgpt4_0.png}
&\includegraphics[width=0.2\textwidth]{fig/ripley0-6_chatgpt4_0.png}
%&\includegraphics[width=0.2\textwidth]{fig/barcode0-6_chatgpt4_0.png}
&\includegraphics[width=0.2\textwidth]{fig/persistence0-6_chatgpt4_0.png}
&\includegraphics[width=0.2\textwidth]{fig/logPersistence0-6_chatgpt4_0.png}\\
0
&\includegraphics[width=0.2\textwidth]{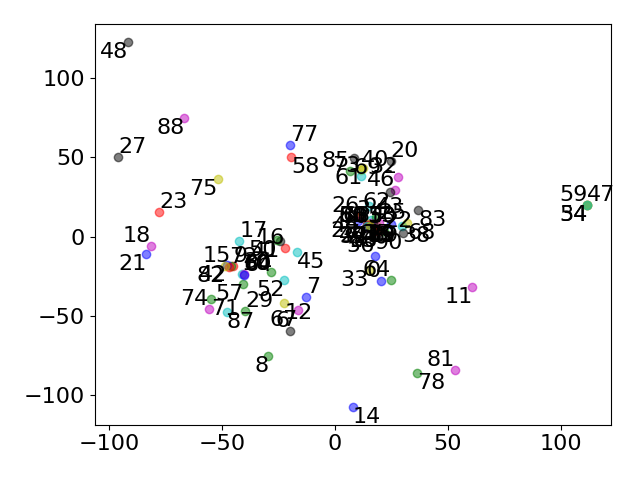}
%&\includegraphics[width=0.2\textwidth]{fig/stress0_chatgpt4_0.png}
&\includegraphics[width=0.2\textwidth]{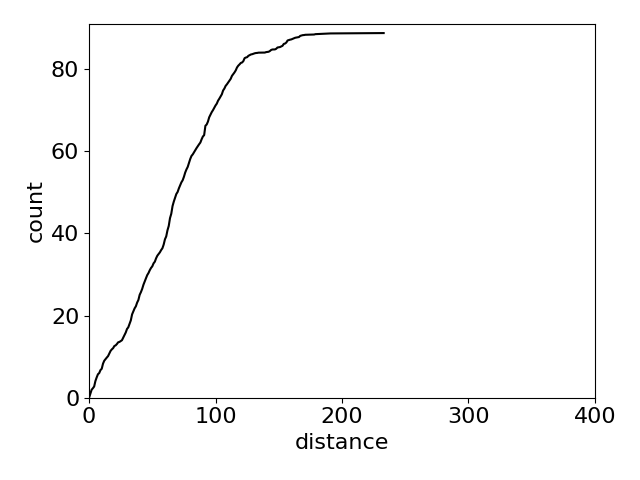}
%&\includegraphics[width=0.2\textwidth]{fig/barcode0_chatgpt4_0.png}
&\includegraphics[width=0.2\textwidth]{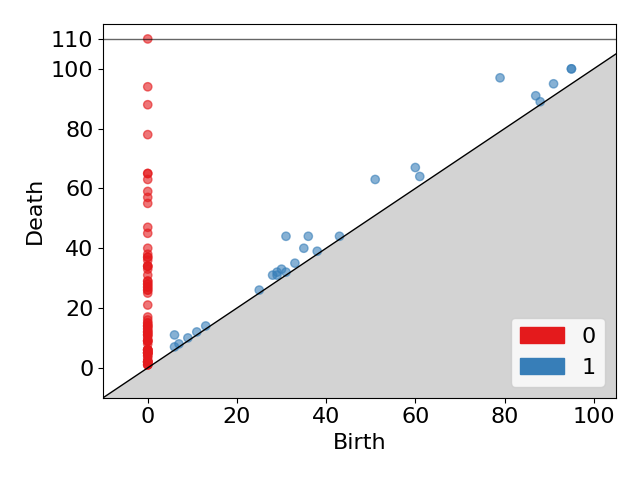}
&\includegraphics[width=0.2\textwidth]{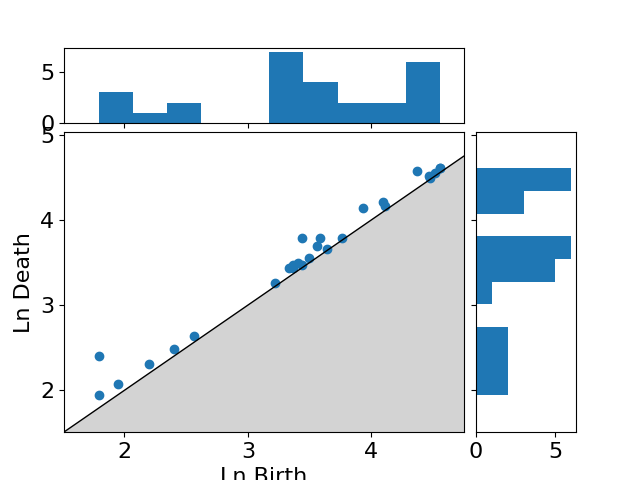}\\
1
&\includegraphics[width=0.2\textwidth]{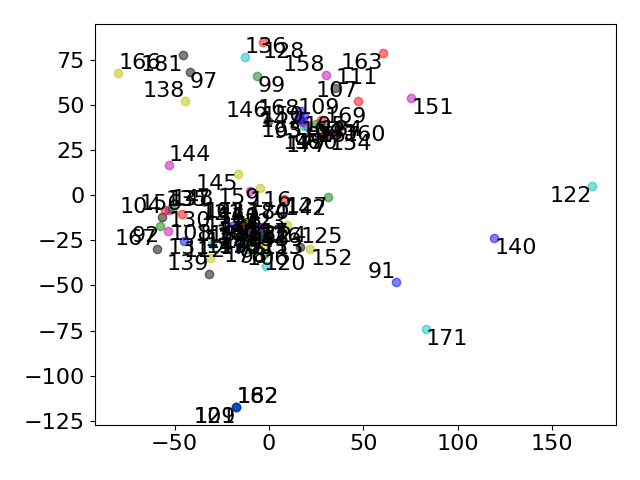}
%&\includegraphics[width=0.2\textwidth]{fig/stress1_chatgpt4_0.png}
&\includegraphics[width=0.2\textwidth]{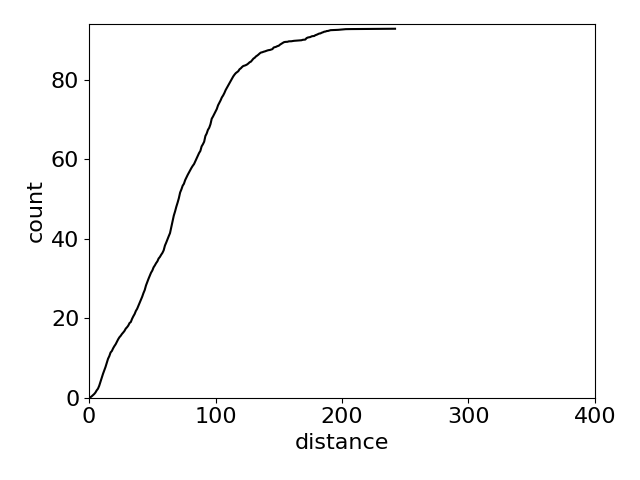}
%&\includegraphics[width=0.2\textwidth]{fig/barcode1_chatgpt4_0.png}
&\includegraphics[width=0.2\textwidth]{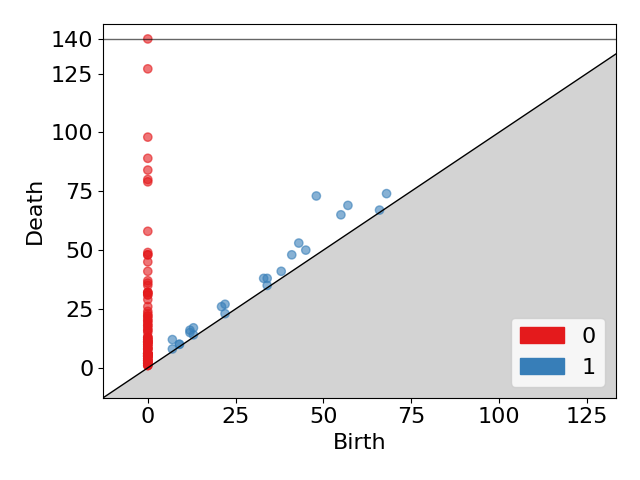}
&\includegraphics[width=0.2\textwidth]{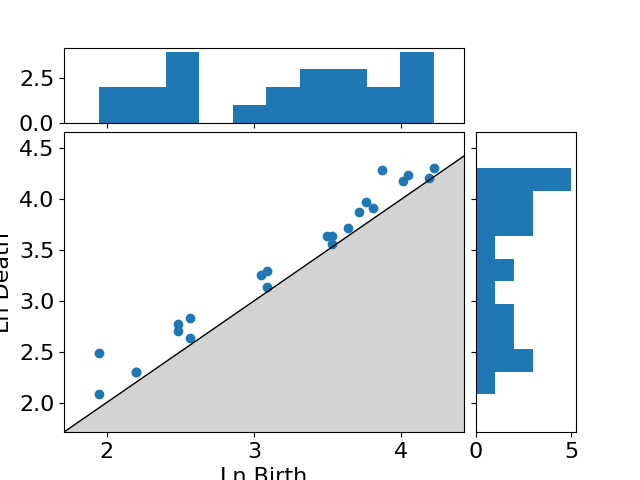}\\
2
&\includegraphics[width=0.2\textwidth]{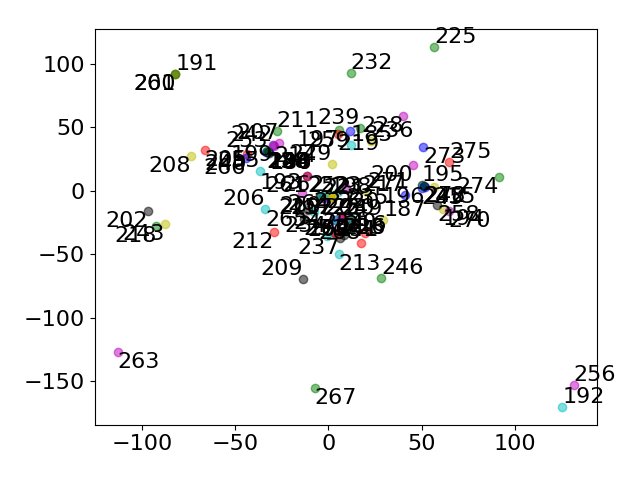}
%&\includegraphics[width=0.2\textwidth]{fig/stress2_chatgpt4_0.png}
&\includegraphics[width=0.2\textwidth]{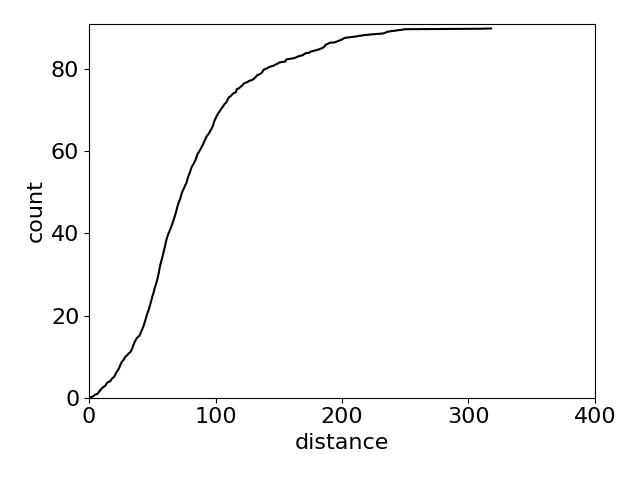}
%&\includegraphics[width=0.2\textwidth]{fig/barcode2_chatgpt4_0.png}
&\includegraphics[width=0.2\textwidth]{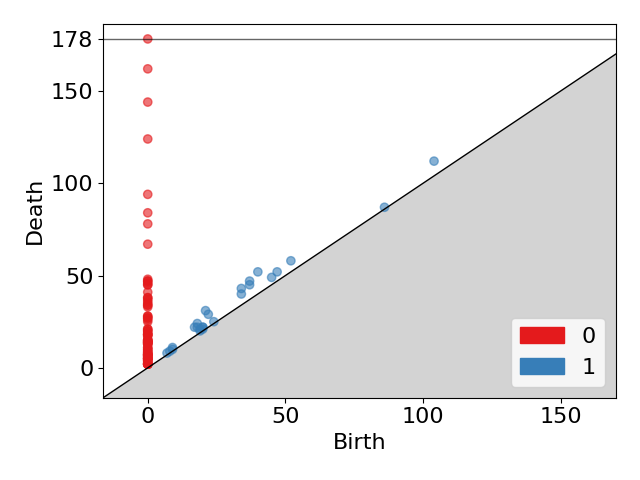}
&\includegraphics[width=0.2\textwidth]{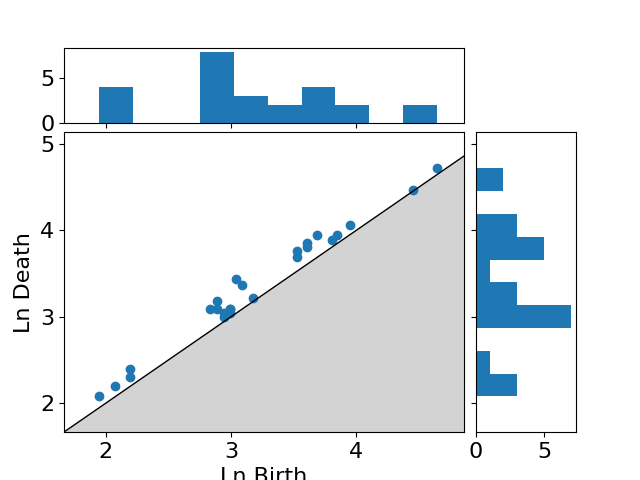}\\
3
&\includegraphics[width=0.2\textwidth]{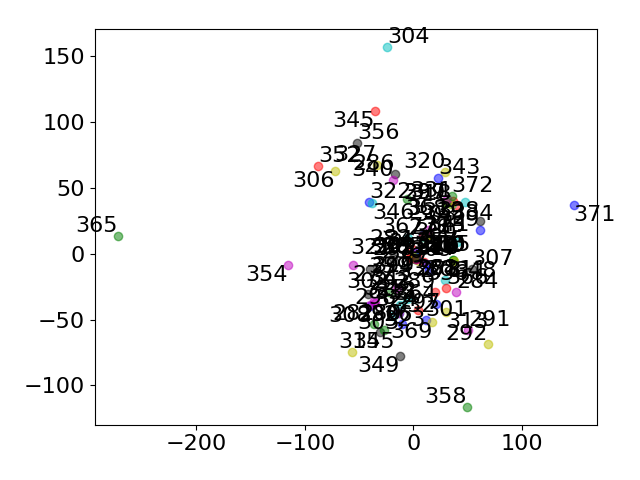}
%&\includegraphics[width=0.2\textwidth]{fig/stress3_chatgpt4_0.png}
&\includegraphics[width=0.2\textwidth]{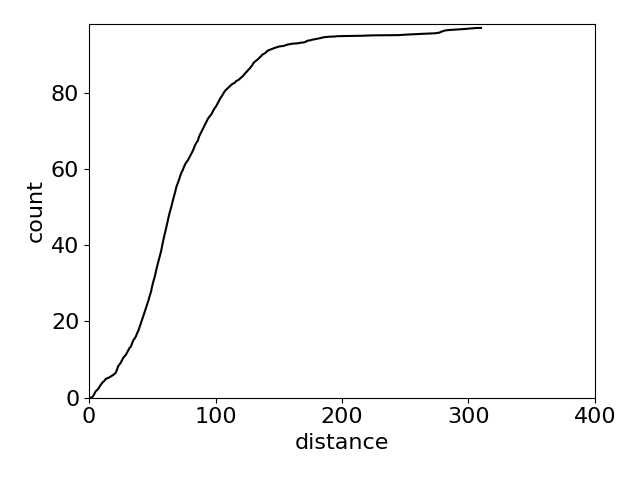}
%&\includegraphics[width=0.2\textwidth]{fig/barcode3_chatgpt4_0.png}
&\includegraphics[width=0.2\textwidth]{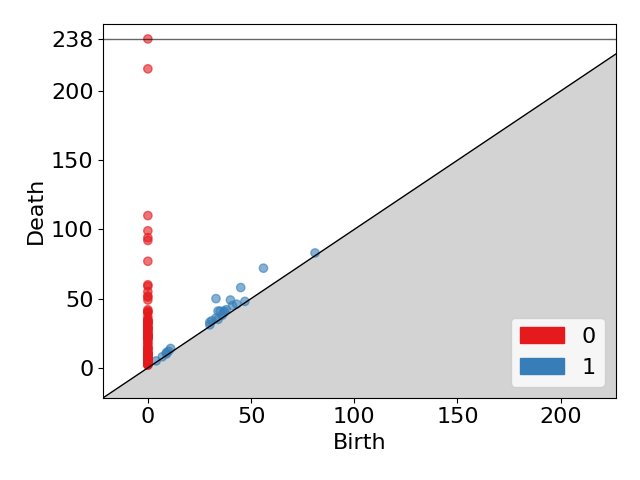}
&\includegraphics[width=0.2\textwidth]{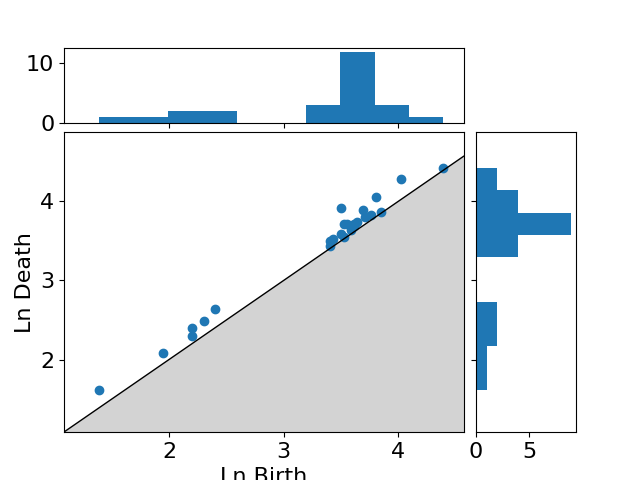}\\
4
&\includegraphics[width=0.2\textwidth]{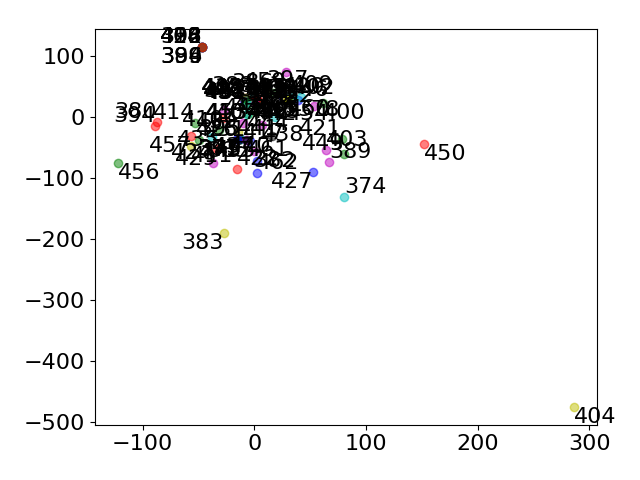}
%&\includegraphics[width=0.2\textwidth]{fig/stress4_chatgpt4_0.png}
&\includegraphics[width=0.2\textwidth]{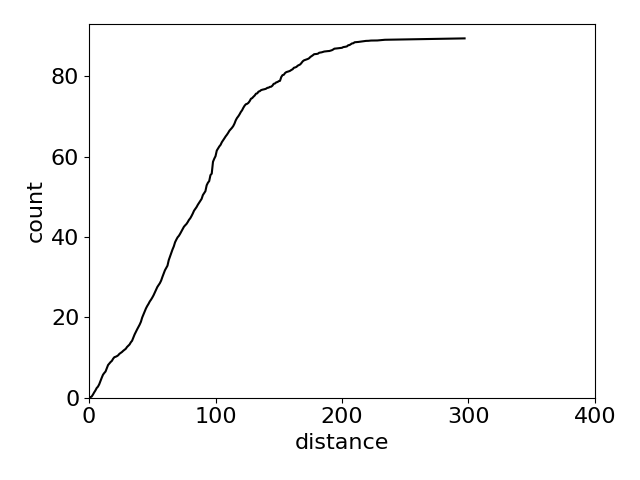}
%&\includegraphics[width=0.2\textwidth]{fig/barcode4_chatgpt4_0.png}
&\includegraphics[width=0.2\textwidth]{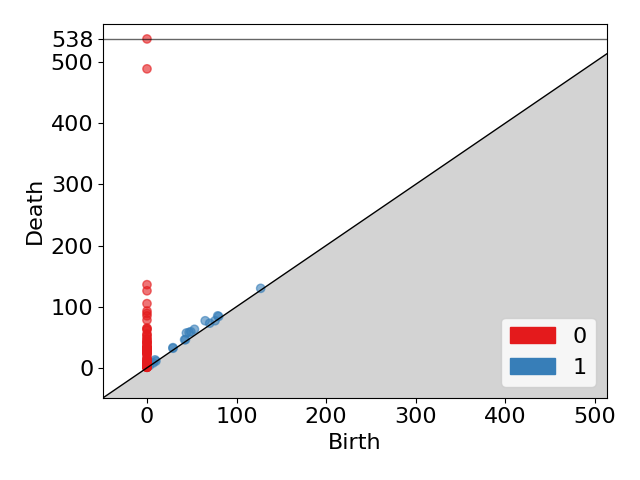}
&\includegraphics[width=0.2\textwidth]{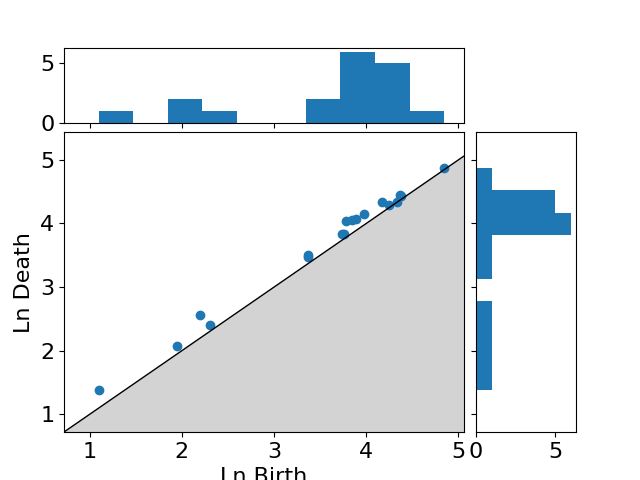}\\
5
&\includegraphics[width=0.2\textwidth]{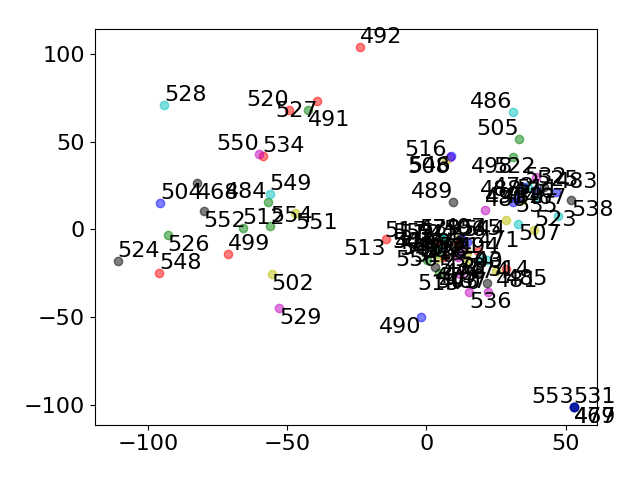}
%&\includegraphics[width=0.2\textwidth]{fig/stress5_chatgpt4_0.png}
&\includegraphics[width=0.2\textwidth]{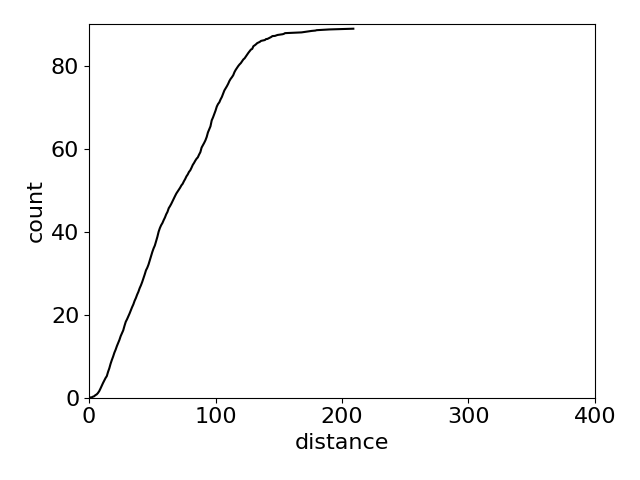}
%&\includegraphics[width=0.2\textwidth]{fig/barcode5_chatgpt4_0.png}
&\includegraphics[width=0.2\textwidth]{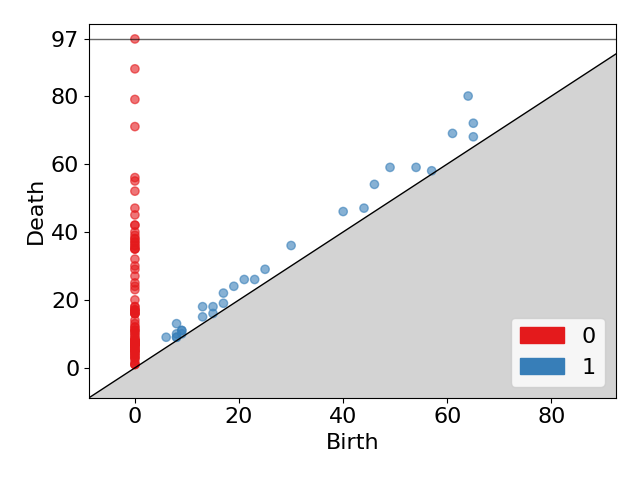}
&\includegraphics[width=0.2\textwidth]{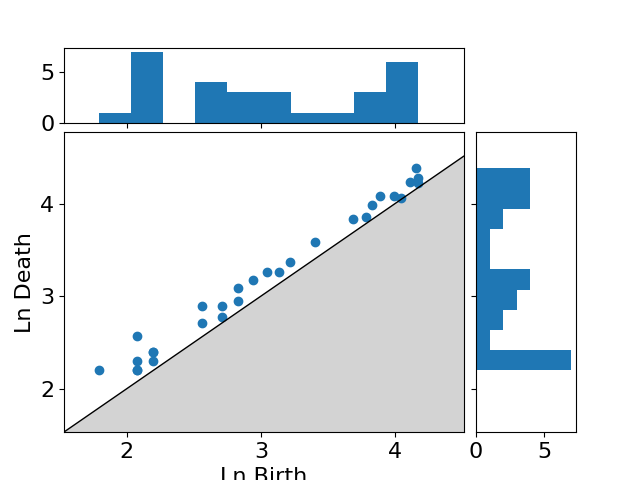}\\
6
&\includegraphics[width=0.2\textwidth]{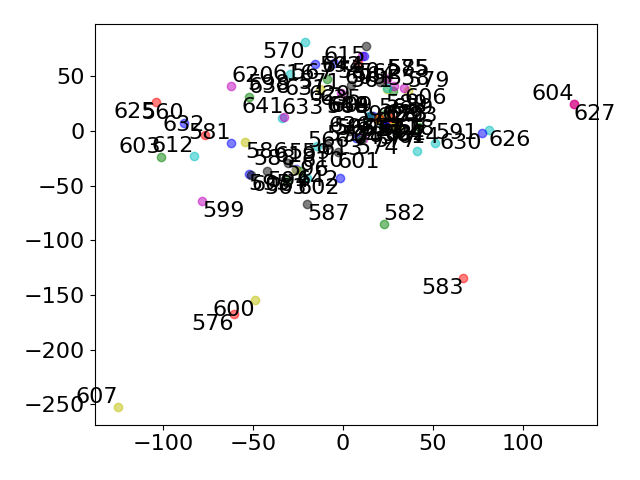}
%&\includegraphics[width=0.2\textwidth]{fig/stress6_chatgpt4_0.png}
&\includegraphics[width=0.2\textwidth]{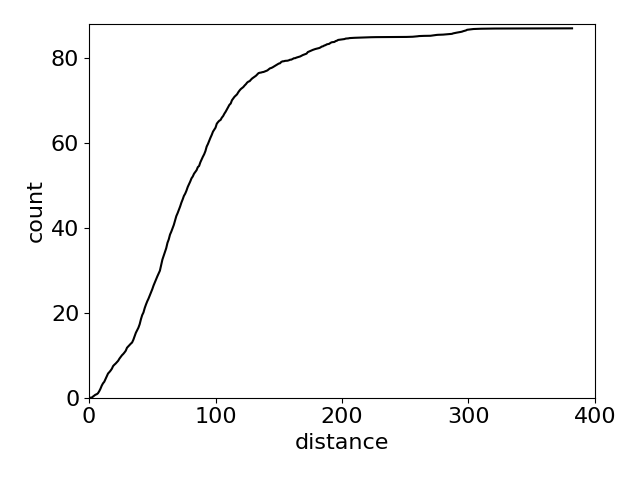}
%&\includegraphics[width=0.2\textwidth]{fig/barcode6_chatgpt4_0.png}
&\includegraphics[width=0.2\textwidth]{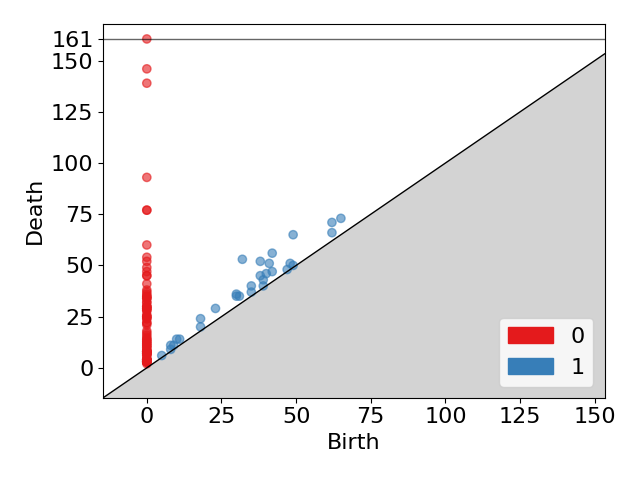}
&\includegraphics[width=0.2\textwidth]{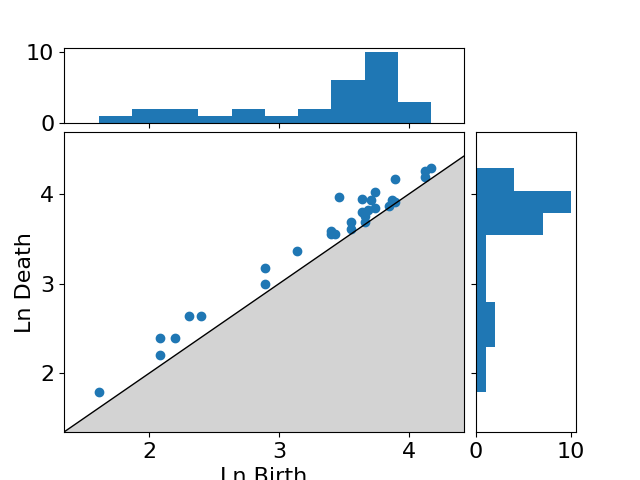}\\
\end{tabular}
\caption{Analysis of the program part of the output of ChatGPT-4.}
\label{fig:chatGPT4}
\end{figure*}

\begin{figure*}
\centering
\begin{tabular}{c|cccc}
Group & Projection & Ripley's K-function & Persistence diagram & Logarithmic Persistence \\\hline
0-6
&\includegraphics[width=0.2\textwidth]{fig/embedding0-6_tinyllama_0.5.png}
%&\includegraphics[width=0.2\textwidth]{fig/stress0-6_tinyllama_0.5.png}
&\includegraphics[width=0.2\textwidth]{fig/ripley0-6_tinyllama_0.5.png}
%&\includegraphics[width=0.2\textwidth]{fig/barcode0-6_tinyllama_0.5.png}
&\includegraphics[width=0.2\textwidth]{fig/persistence0-6_tinyllama_0.5.png}
&\includegraphics[width=0.2\textwidth]{fig/logPersistence0-6_tinyllama_0.5.png}\\
0
&\includegraphics[width=0.2\textwidth]{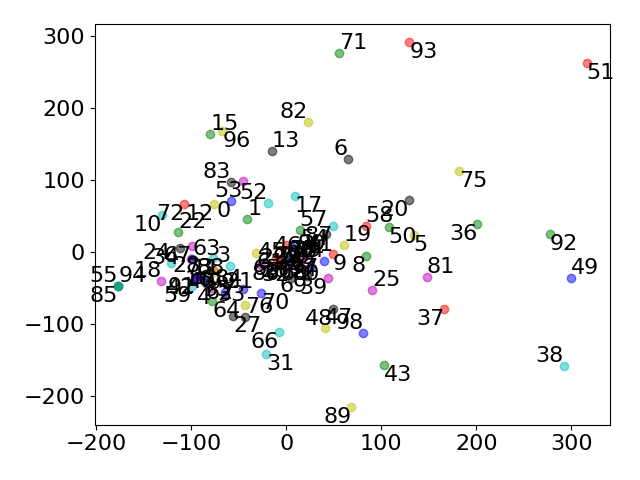}
%&\includegraphics[width=0.2\textwidth]{fig/stress0_tinyllama_0.5.png}
&\includegraphics[width=0.2\textwidth]{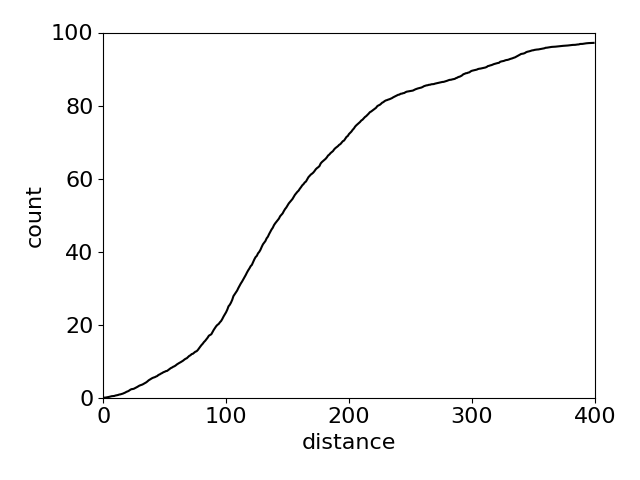}
%&\includegraphics[width=0.2\textwidth]{fig/barcode0_tinyllama_0.5.png}
&\includegraphics[width=0.2\textwidth]{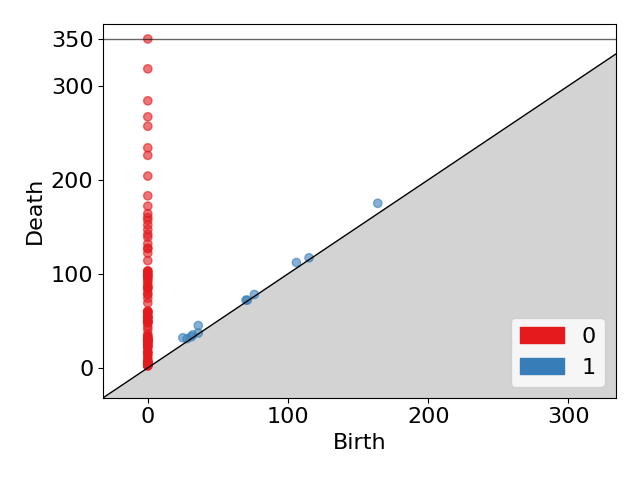}
&\includegraphics[width=0.2\textwidth]{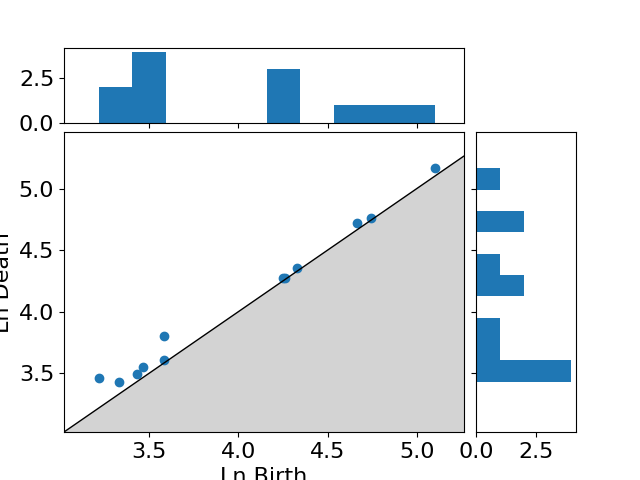}\\
1
&\includegraphics[width=0.2\textwidth]{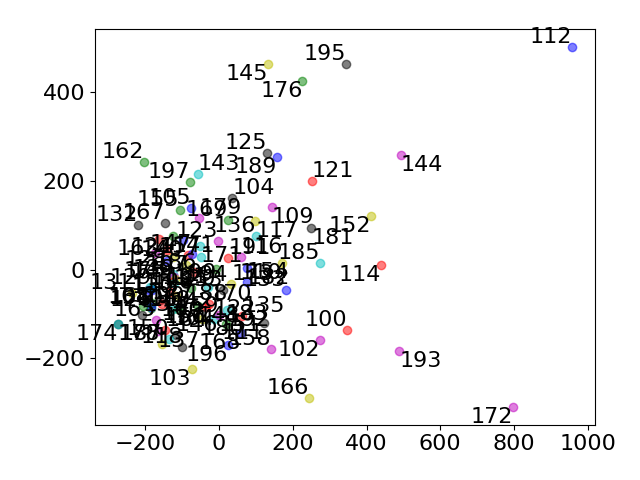}
%&\includegraphics[width=0.2\textwidth]{fig/stress1_tinyllama_0.5.png}
&\includegraphics[width=0.2\textwidth]{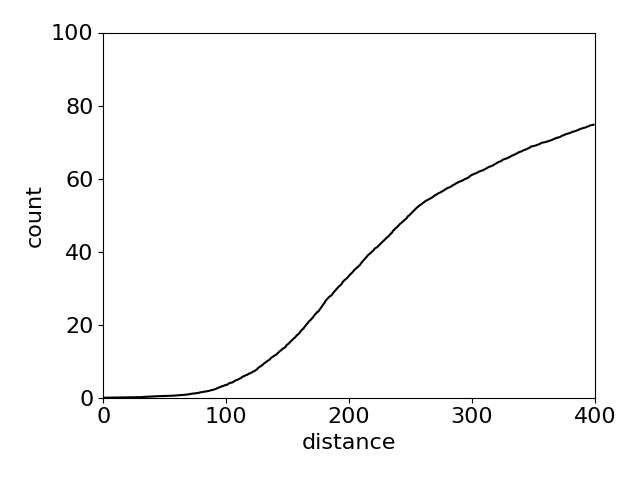}
%&\includegraphics[width=0.2\textwidth]{fig/barcode1_tinyllama_0.5.png}
&\includegraphics[width=0.2\textwidth]{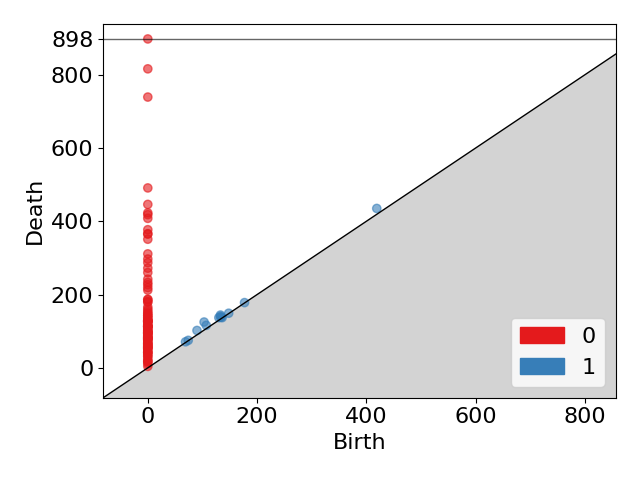}
&\includegraphics[width=0.2\textwidth]{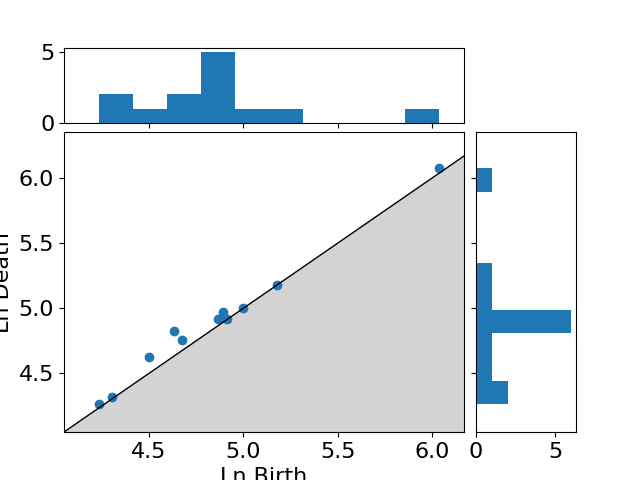}\\
2
&\includegraphics[width=0.2\textwidth]{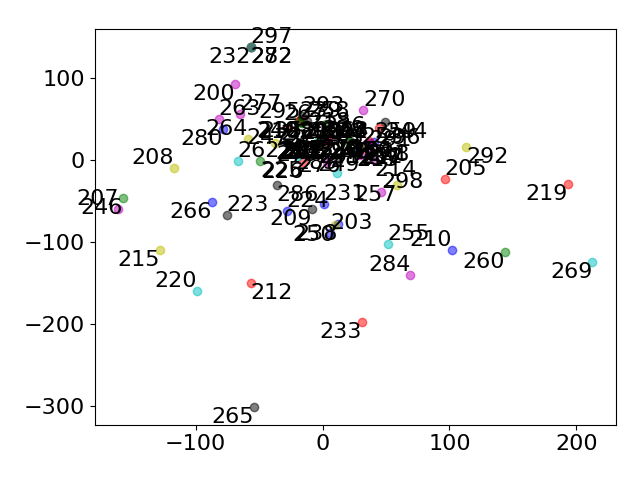}
%&\includegraphics[width=0.2\textwidth]{fig/stress2_tinyllama_0.5.png}
&\includegraphics[width=0.2\textwidth]{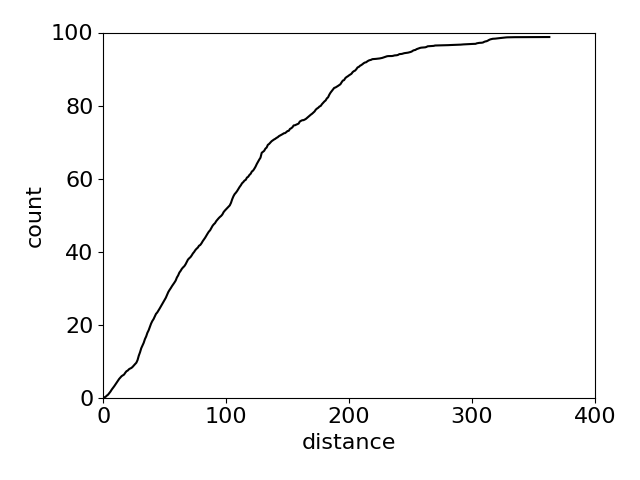}
%&\includegraphics[width=0.2\textwidth]{fig/barcode2_tinyllama_0.5.png}
&\includegraphics[width=0.2\textwidth]{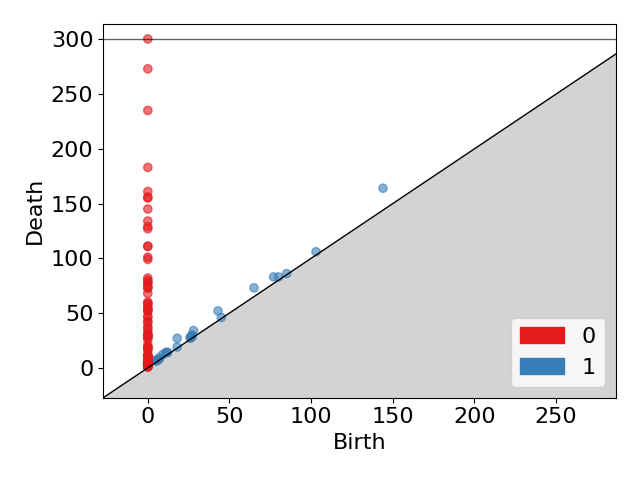}
&\includegraphics[width=0.2\textwidth]{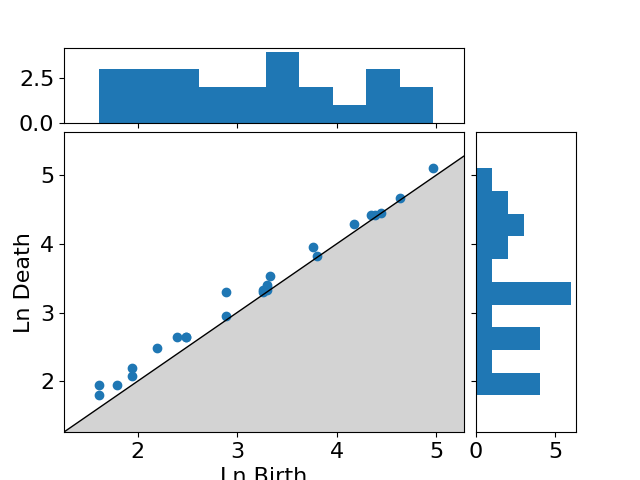}\\
3
&\includegraphics[width=0.2\textwidth]{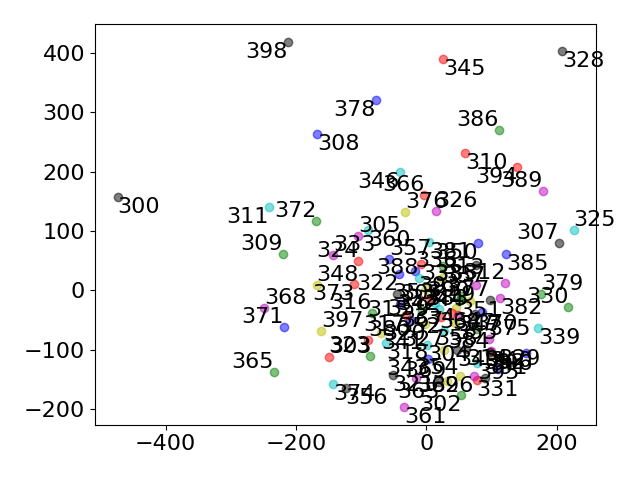}
%&\includegraphics[width=0.2\textwidth]{fig/stress3_tinyllama_0.5.png}
&\includegraphics[width=0.2\textwidth]{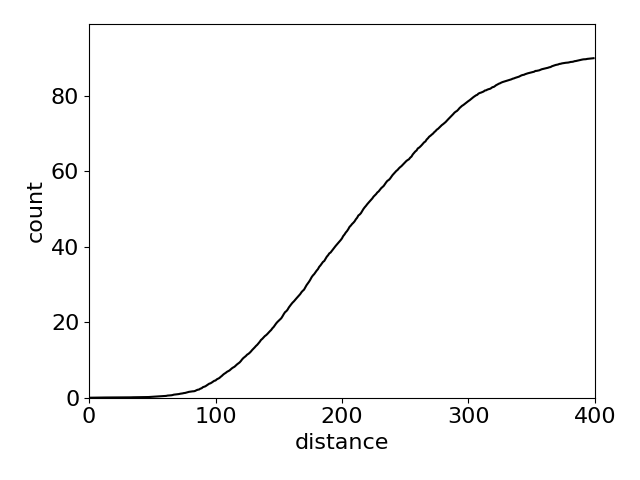}
%&\includegraphics[width=0.2\textwidth]{fig/barcode3_tinyllama_0.5.png}
&\includegraphics[width=0.2\textwidth]{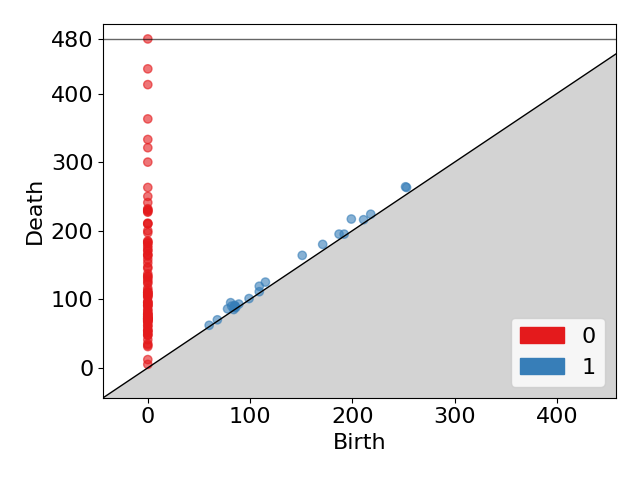}
&\includegraphics[width=0.2\textwidth]{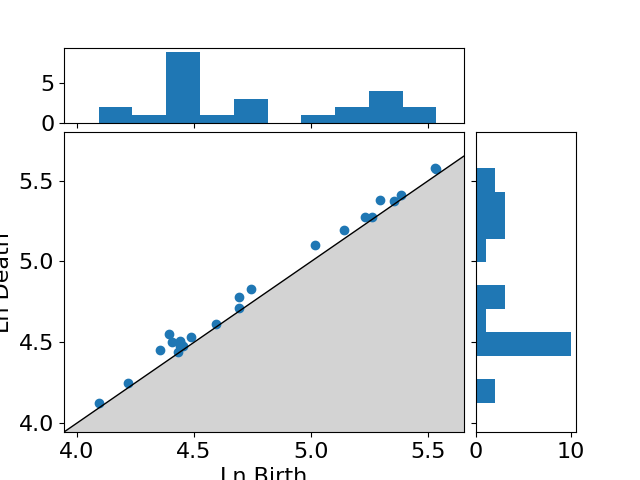}\\
4
&\includegraphics[width=0.2\textwidth]{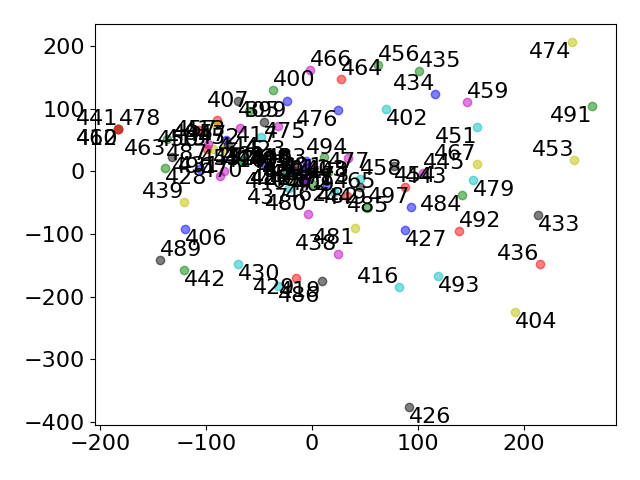}
%&\includegraphics[width=0.2\textwidth]{fig/stress4_tinyllama_0.5.png}
&\includegraphics[width=0.2\textwidth]{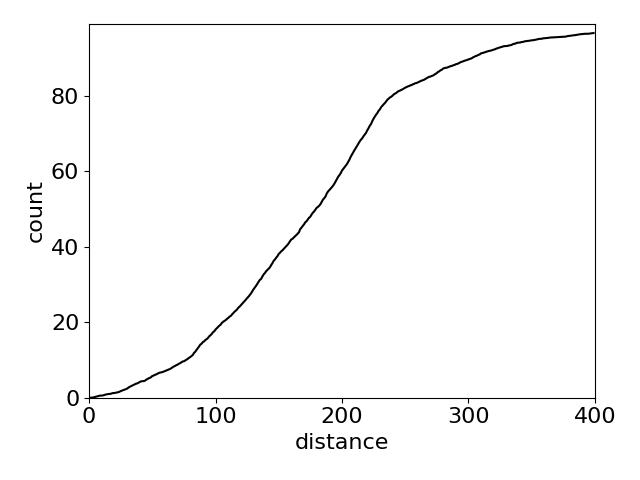}
%&\includegraphics[width=0.2\textwidth]{fig/barcode4_tinyllama_0.5.png}
&\includegraphics[width=0.2\textwidth]{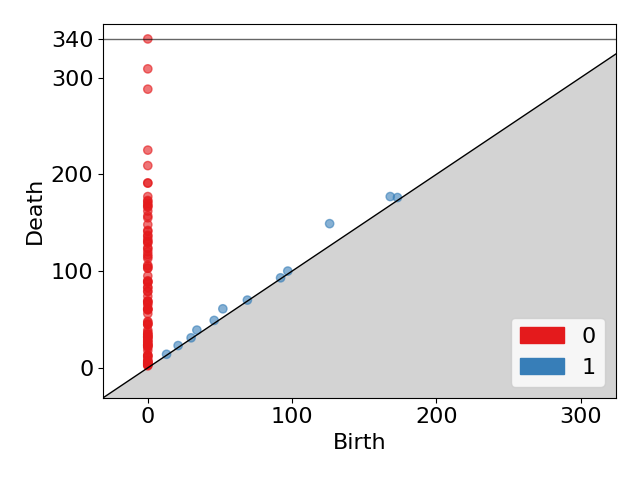}
&\includegraphics[width=0.2\textwidth]{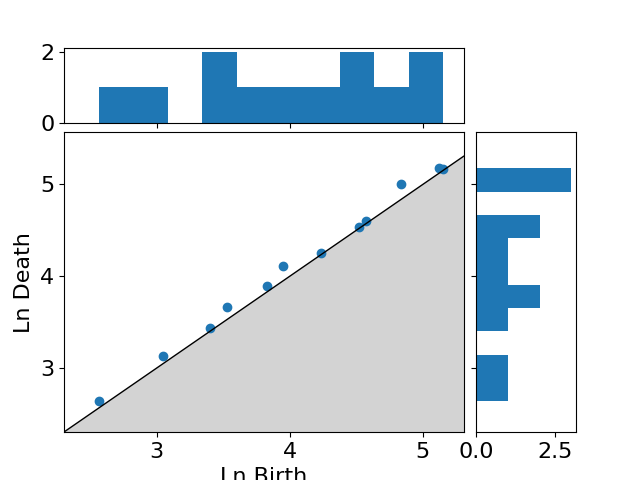}\\
5
&\includegraphics[width=0.2\textwidth]{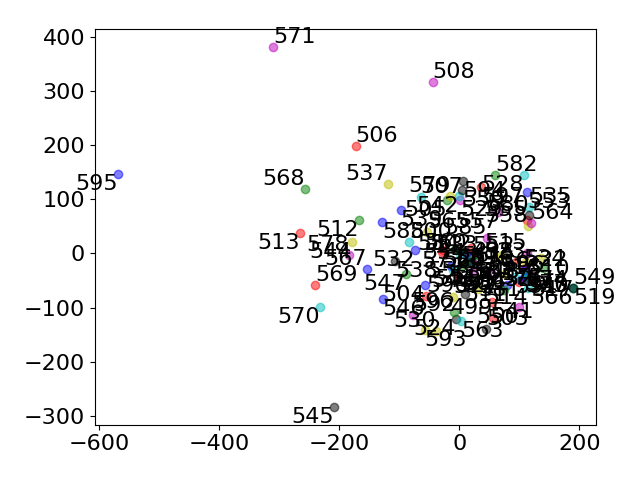}
%&\includegraphics[width=0.2\textwidth]{fig/stress5_tinyllama_0.5.png}
&\includegraphics[width=0.2\textwidth]{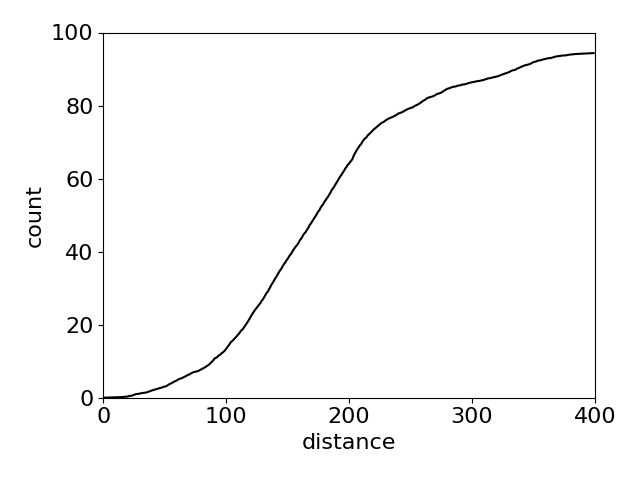}
%&\includegraphics[width=0.2\textwidth]{fig/barcode5_tinyllama_0.5.png}
&\includegraphics[width=0.2\textwidth]{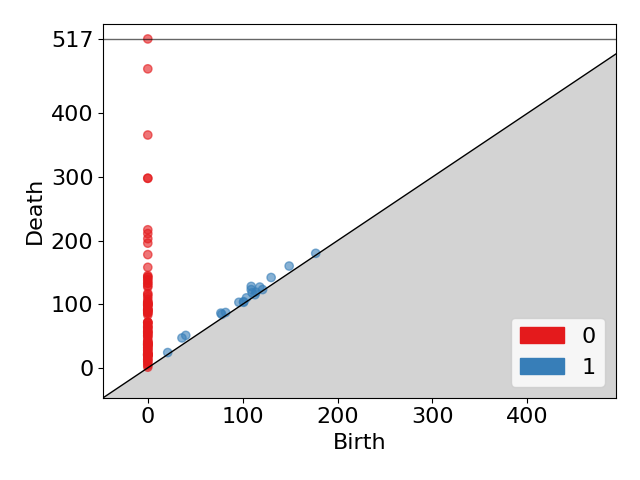}
&\includegraphics[width=0.2\textwidth]{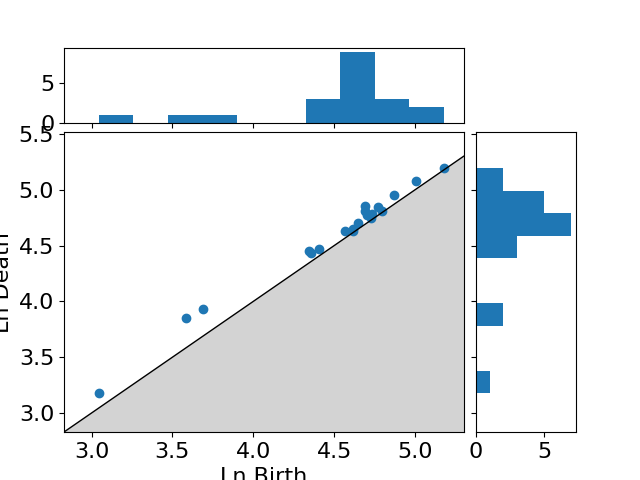}\\
6
&\includegraphics[width=0.2\textwidth]{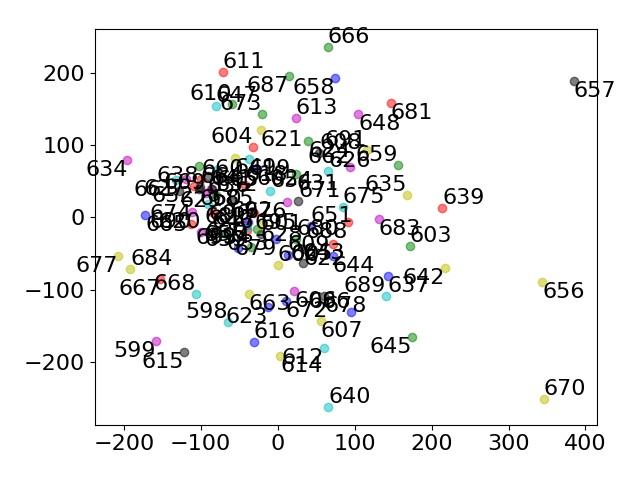}
%&\includegraphics[width=0.2\textwidth]{fig/stress6_tinyllama_0.5.png}
&\includegraphics[width=0.2\textwidth]{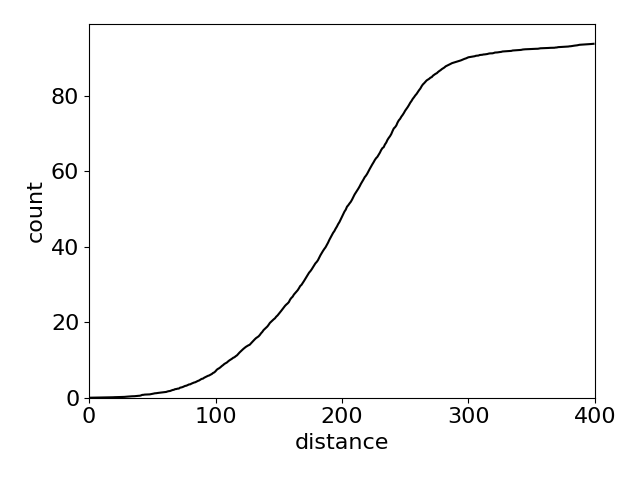}
%&\includegraphics[width=0.2\textwidth]{fig/barcode6_tinyllama_0.5.png}
&\includegraphics[width=0.2\textwidth]{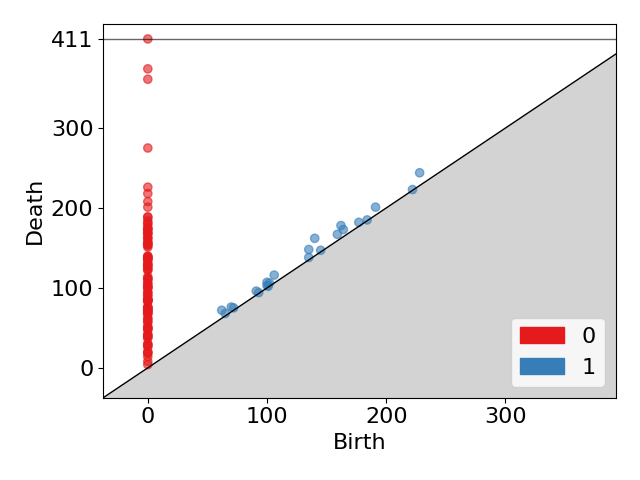}
&\includegraphics[width=0.2\textwidth]{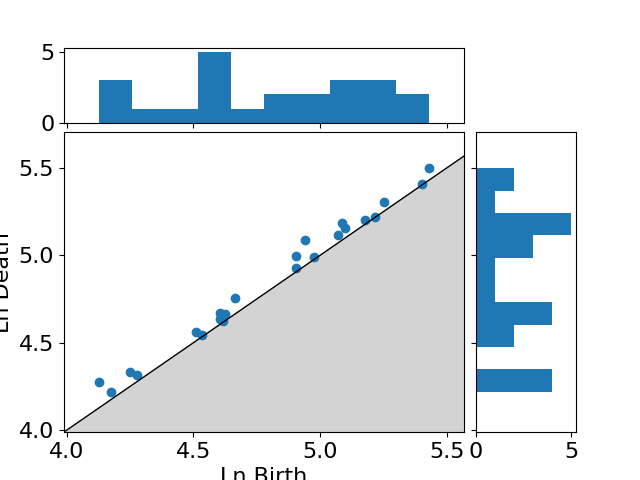}\\
\end{tabular}
\caption{Analysis of the program part of the output of TinyLlama with $t=0.5$.}
\label{fig:tinyllama05}
\end{figure*}
\begin{figure*}
\centering
\begin{tabular}{c|cccc}
Group & Projection & Ripley's K-function & Persistence diagram & Logarithmic Persistence \\\hline
0-6
&\includegraphics[width=0.2\textwidth]{fig/embedding0-6_tinyllama_0.7.png}
%&\includegraphics[width=0.2\textwidth]{fig/stress0-6_tinyllama_0.7.png}
&\includegraphics[width=0.2\textwidth]{fig/ripley0-6_tinyllama_0.7.png}
%&\includegraphics[width=0.2\textwidth]{fig/barcode0-6_tinyllama_0.7.png}
&\includegraphics[width=0.2\textwidth]{fig/persistence0-6_tinyllama_0.7.png}
&\includegraphics[width=0.2\textwidth]{fig/logPersistence0-6_tinyllama_0.7.png}\\
0
&\includegraphics[width=0.2\textwidth]{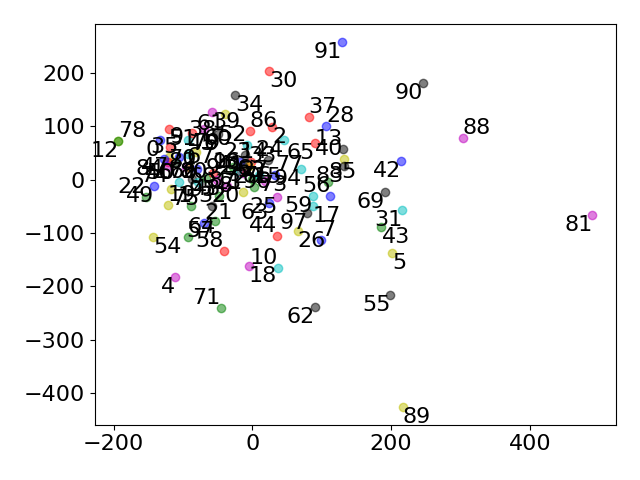}
%&\includegraphics[width=0.2\textwidth]{fig/stress0_tinyllama_0.7.png}
&\includegraphics[width=0.2\textwidth]{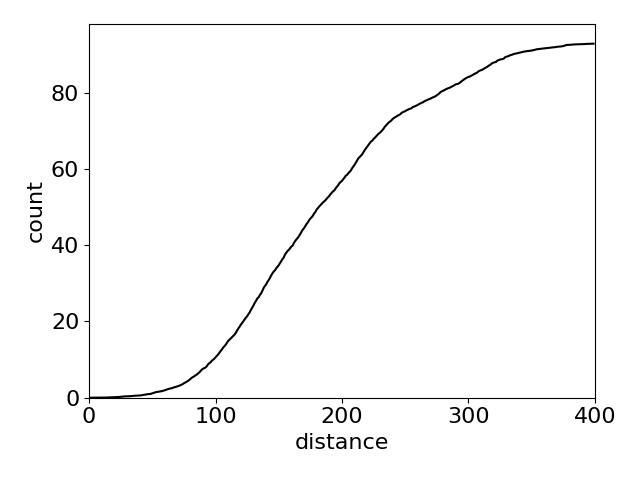}
%&\includegraphics[width=0.2\textwidth]{fig/barcode0_tinyllama_0.7.png}
&\includegraphics[width=0.2\textwidth]{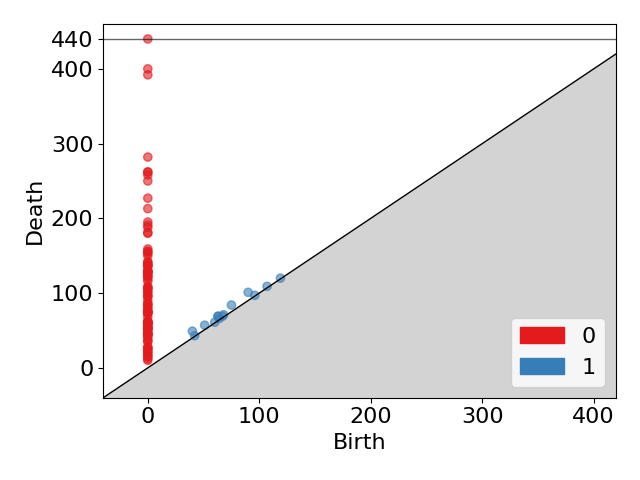}
&\includegraphics[width=0.2\textwidth]{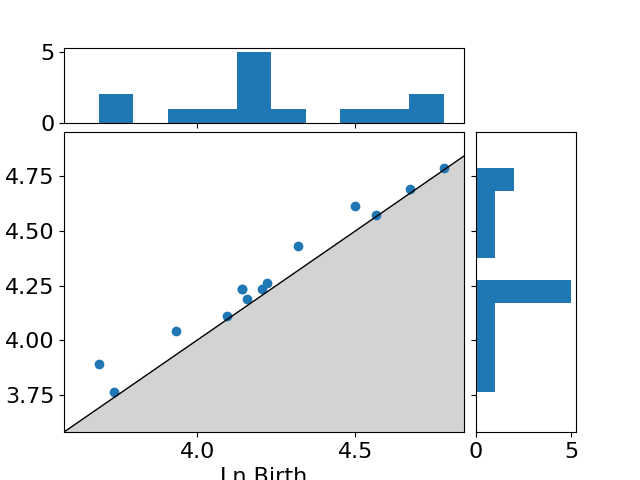}\\
1
&\includegraphics[width=0.2\textwidth]{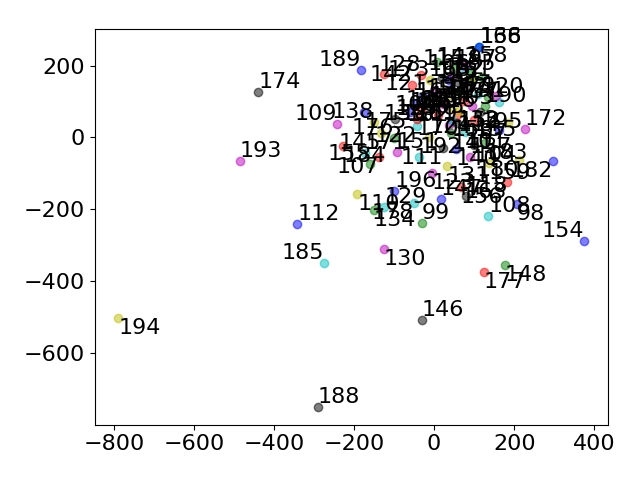}
%&\includegraphics[width=0.2\textwidth]{fig/stress1_tinyllama_0.7.png}
&\includegraphics[width=0.2\textwidth]{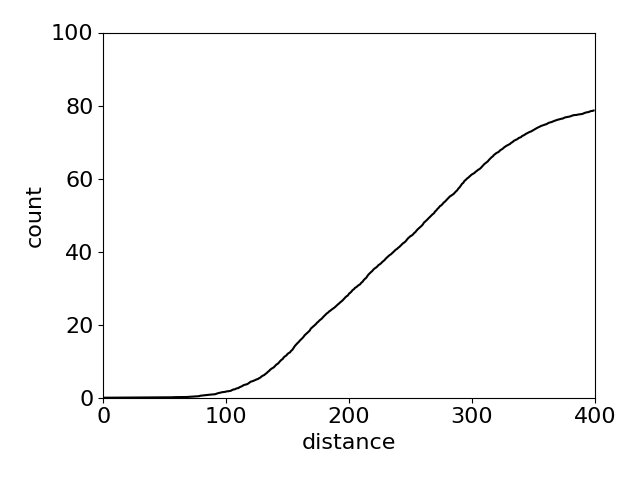}
%&\includegraphics[width=0.2\textwidth]{fig/barcode1_tinyllama_0.7.png}
&\includegraphics[width=0.2\textwidth]{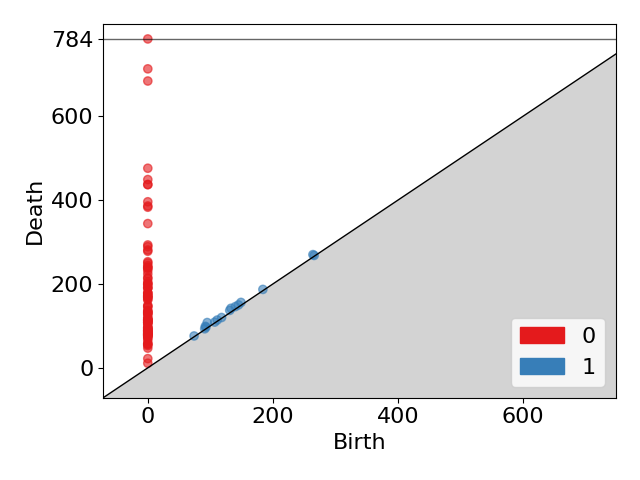}
&\includegraphics[width=0.2\textwidth]{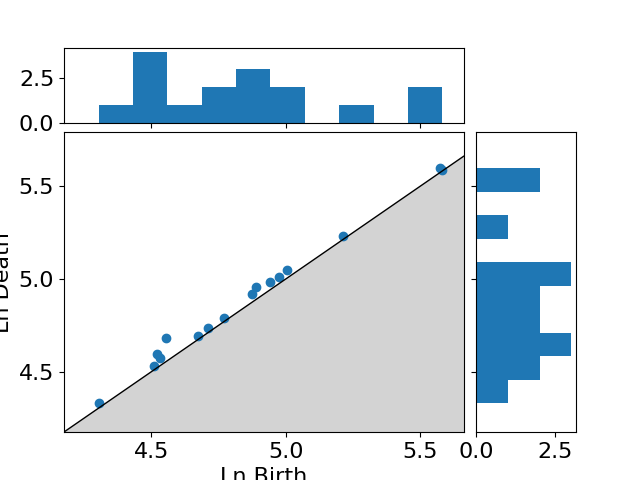}\\
2
&\includegraphics[width=0.2\textwidth]{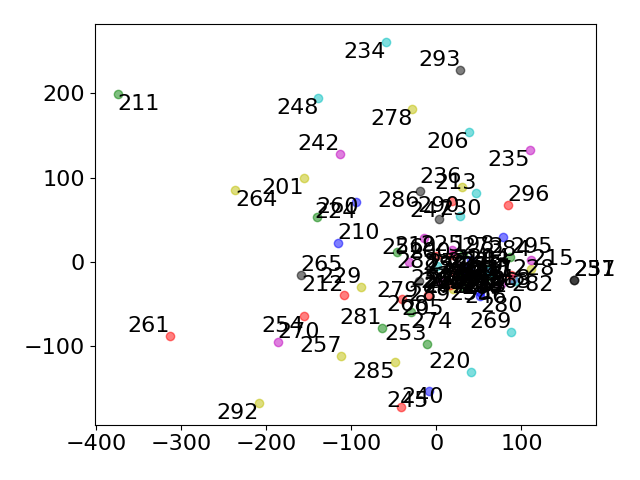}
%&\includegraphics[width=0.2\textwidth]{fig/stress2_tinyllama_0.7.png}
&\includegraphics[width=0.2\textwidth]{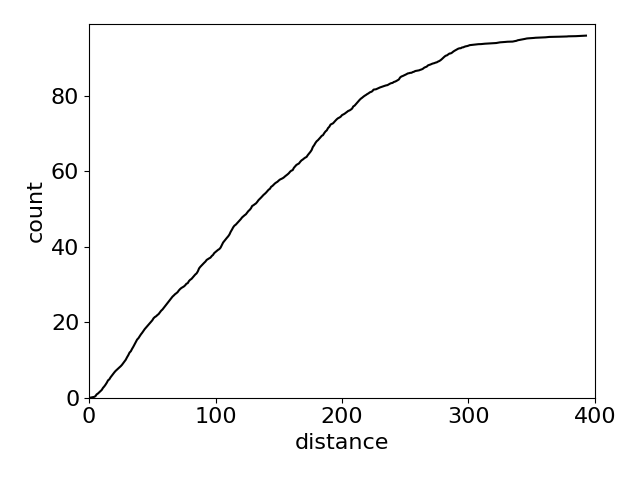}
%&\includegraphics[width=0.2\textwidth]{fig/barcode2_tinyllama_0.7.png}
&\includegraphics[width=0.2\textwidth]{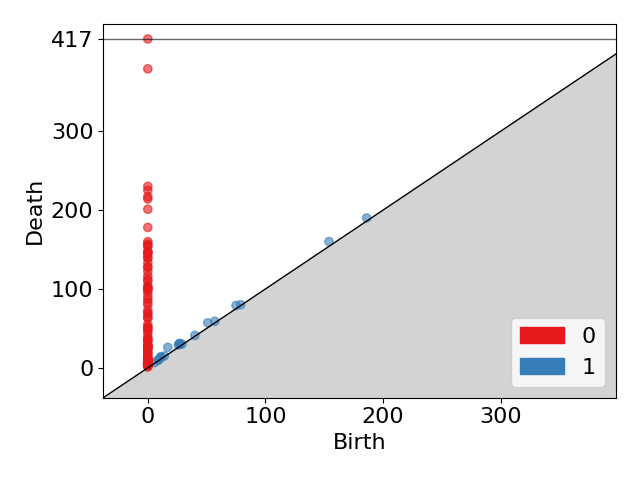}
&\includegraphics[width=0.2\textwidth]{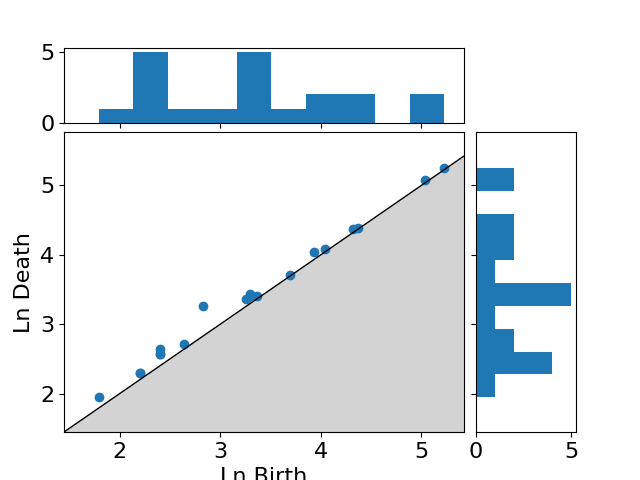}\\
3
&\includegraphics[width=0.2\textwidth]{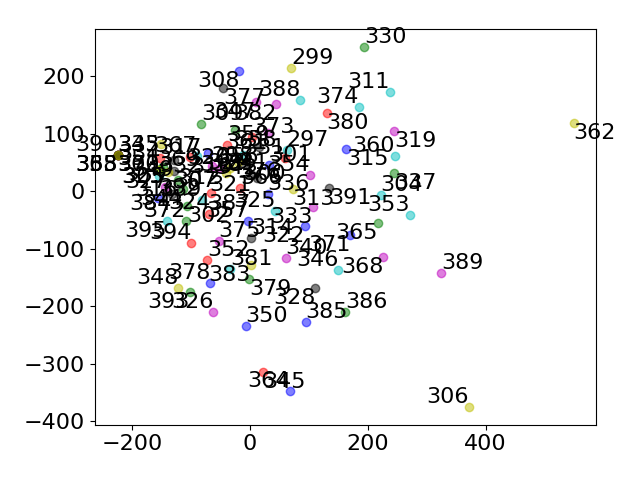}
%&\includegraphics[width=0.2\textwidth]{fig/stress3_tinyllama_0.7.png}
&\includegraphics[width=0.2\textwidth]{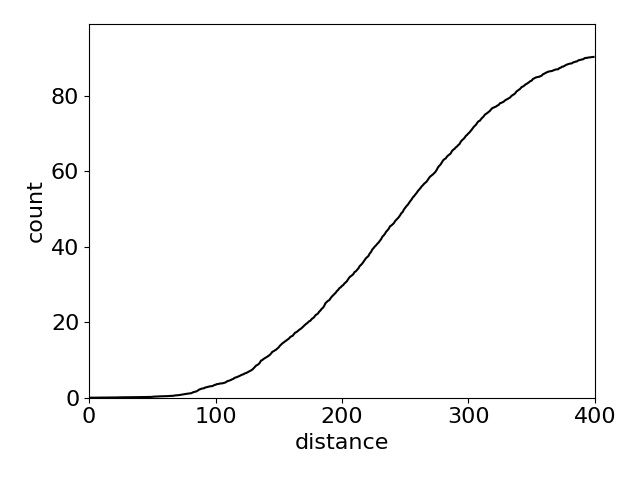}
%&\includegraphics[width=0.2\textwidth]{fig/barcode3_tinyllama_0.7.png}
&\includegraphics[width=0.2\textwidth]{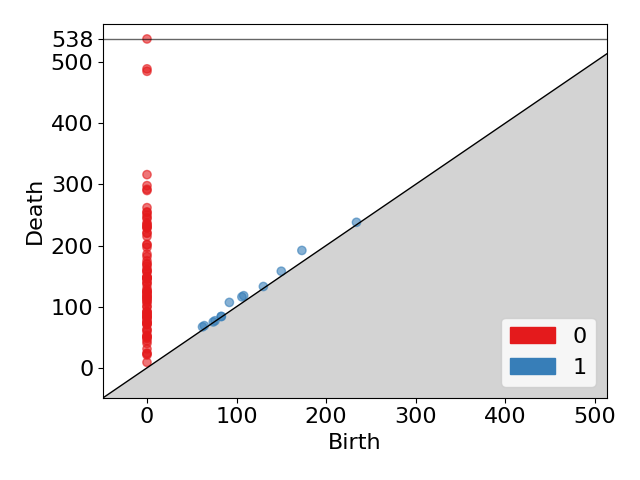}
&\includegraphics[width=0.2\textwidth]{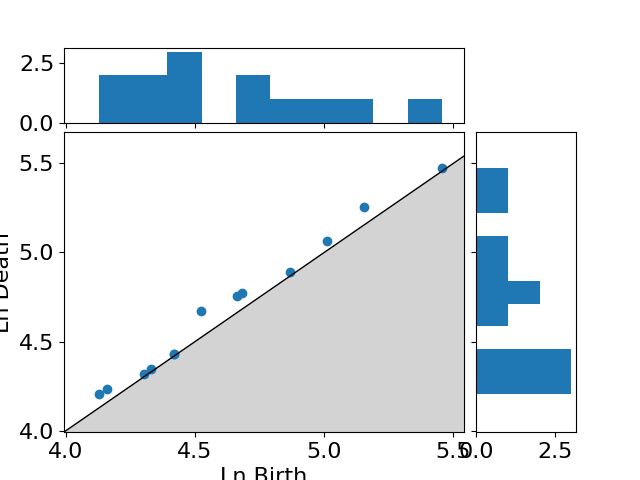}\\
4
&\includegraphics[width=0.2\textwidth]{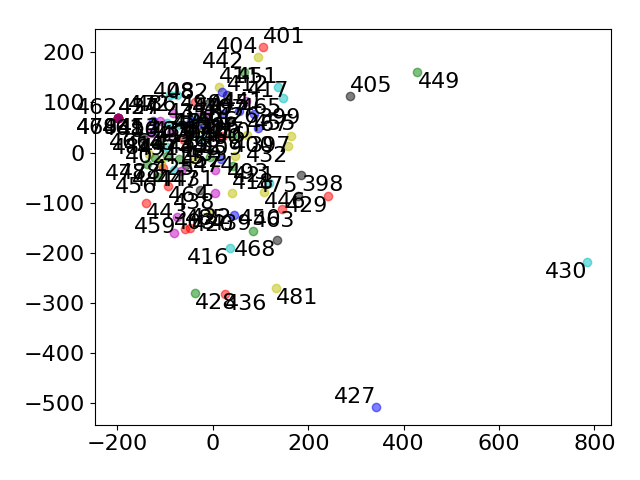}
%&\includegraphics[width=0.2\textwidth]{fig/stress4_tinyllama_0.7.png}
&\includegraphics[width=0.2\textwidth]{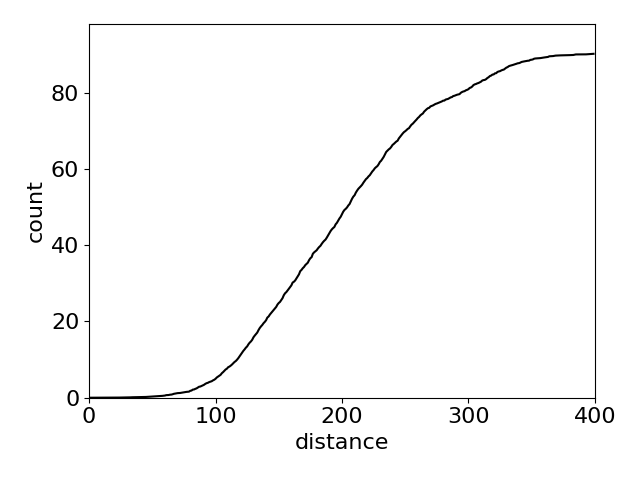}
%&\includegraphics[width=0.2\textwidth]{fig/barcode4_tinyllama_0.7.png}
&\includegraphics[width=0.2\textwidth]{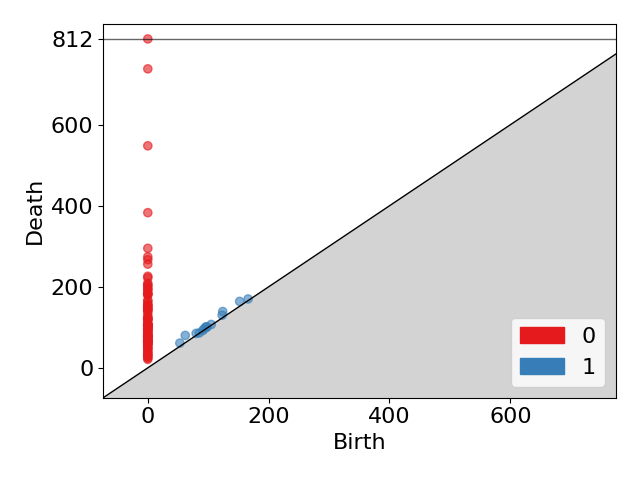}
&\includegraphics[width=0.2\textwidth]{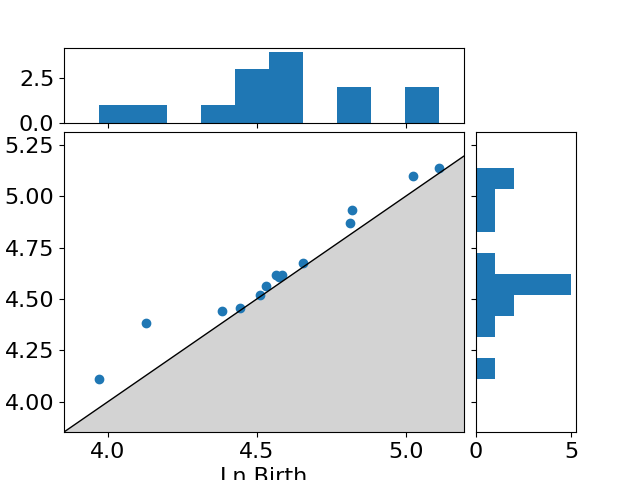}\\
5
&\includegraphics[width=0.2\textwidth]{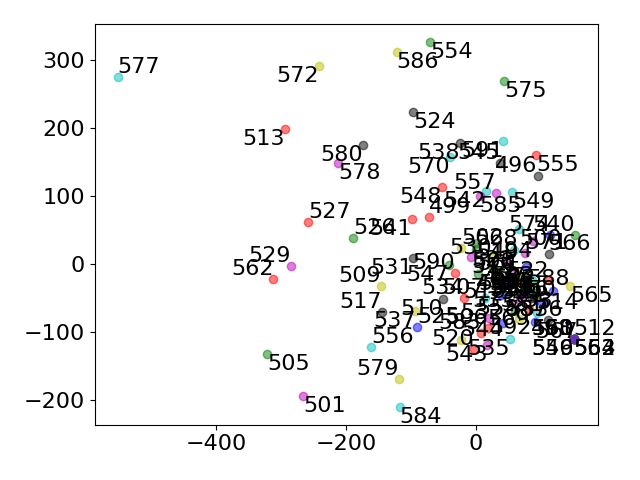}
%&\includegraphics[width=0.2\textwidth]{fig/stress5_tinyllama_0.7.png}
&\includegraphics[width=0.2\textwidth]{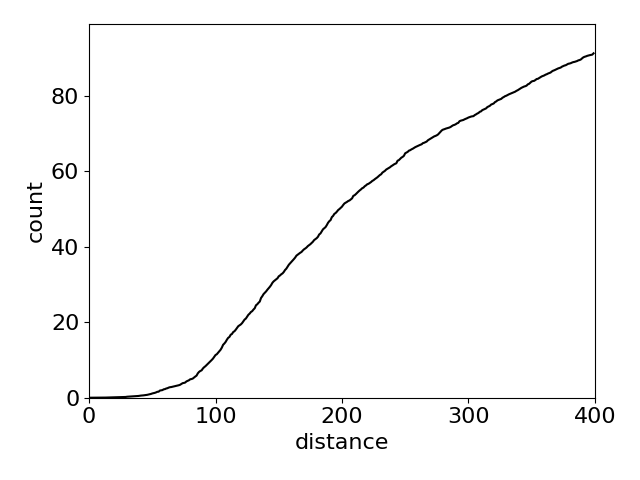}
%&\includegraphics[width=0.2\textwidth]{fig/barcode5_tinyllama_0.7.png}
&\includegraphics[width=0.2\textwidth]{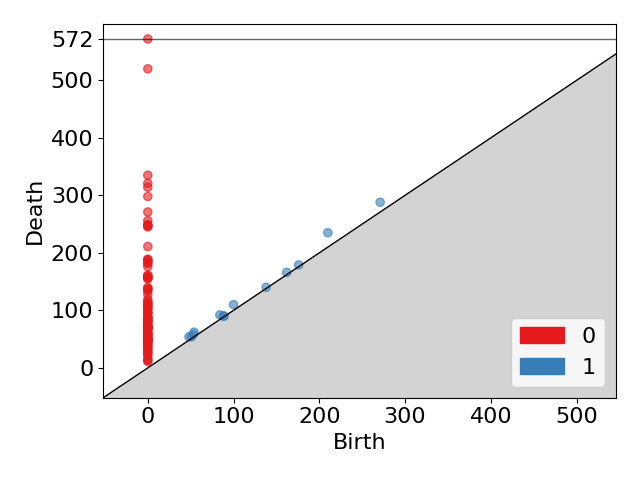}
&\includegraphics[width=0.2\textwidth]{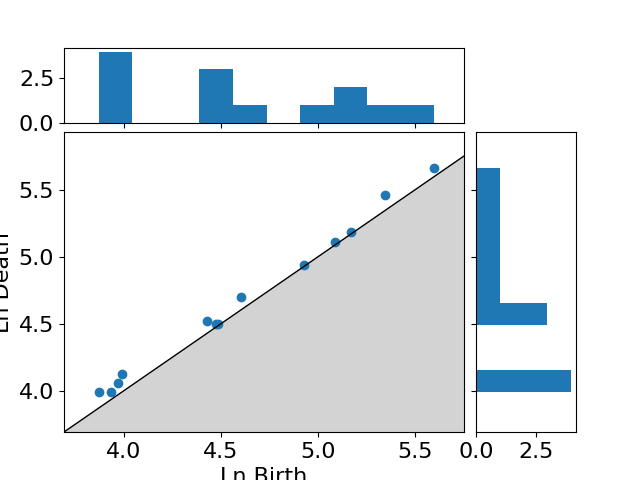}\\
6
&\includegraphics[width=0.2\textwidth]{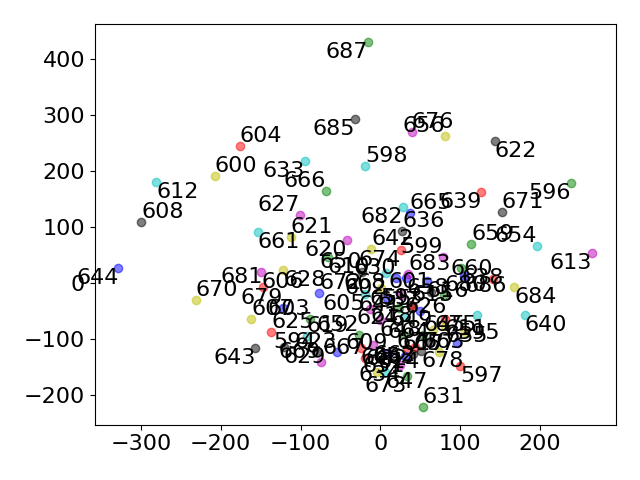}
%&\includegraphics[width=0.2\textwidth]{fig/stress6_tinyllama_0.7.png}
&\includegraphics[width=0.2\textwidth]{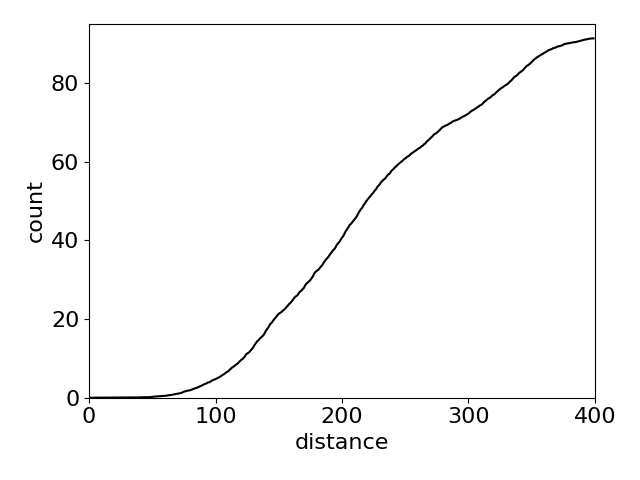}
%&\includegraphics[width=0.2\textwidth]{fig/barcode6_tinyllama_0.7.png}
&\includegraphics[width=0.2\textwidth]{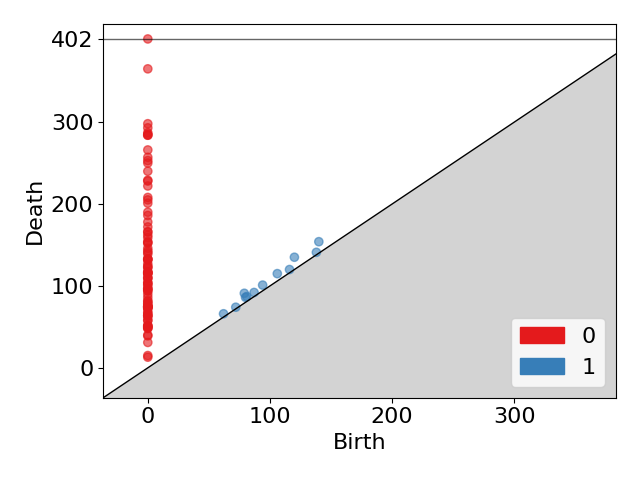}
&\includegraphics[width=0.2\textwidth]{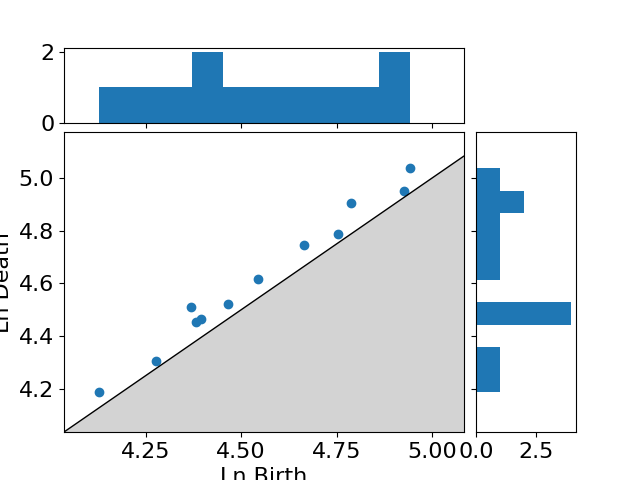}\\
\end{tabular}
\caption{Analysis of the program part of the output of TinyLlama with $t=0.7$}
\label{fig:tinyllama07}
\end{figure*}
\begin{figure*}
\centering
1\begin{tabular}{c|cccc}
Group & Projection & Ripley's K-function & Persistence diagram & Logarithmic Persistence \\\hline
0-6
&\includegraphics[width=0.2\textwidth]{fig/embedding0-6_tinyllama_0.9.png}
%&\includegraphics[width=0.2\textwidth]{fig/stress0-6_tinyllama_0.9.png}
&\includegraphics[width=0.2\textwidth]{fig/ripley0-6_tinyllama_0.9.png}
%&\includegraphics[width=0.2\textwidth]{fig/barcode0-6_tinyllama_0.9.png}
&\includegraphics[width=0.2\textwidth]{fig/persistence0-6_tinyllama_0.9.png}
&\includegraphics[width=0.2\textwidth]{fig/logPersistence0-6_tinyllama_0.9.png}\\
0
&\includegraphics[width=0.2\textwidth]{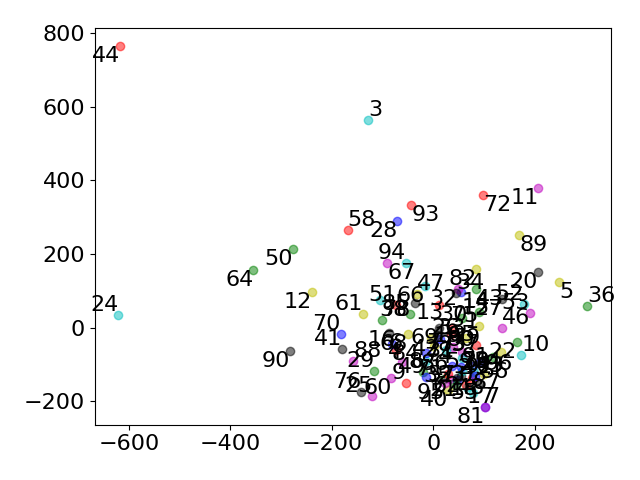}
%&\includegraphics[width=0.2\textwidth]{fig/stress0_tinyllama_0.9.png}
&\includegraphics[width=0.2\textwidth]{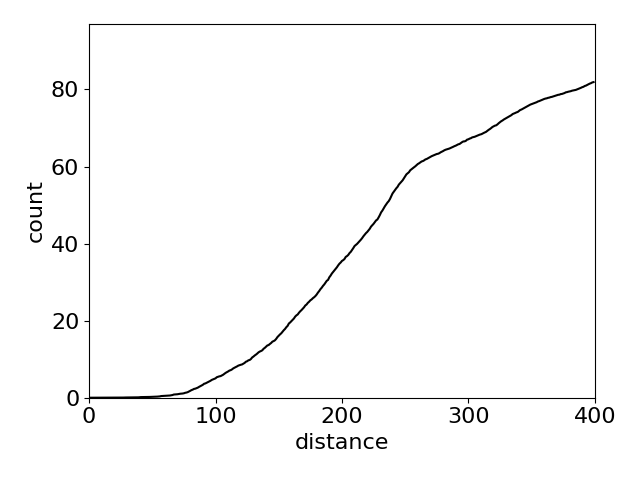}
%&\includegraphics[width=0.2\textwidth]{fig/barcode0_tinyllama_0.9.png}
&\includegraphics[width=0.2\textwidth]{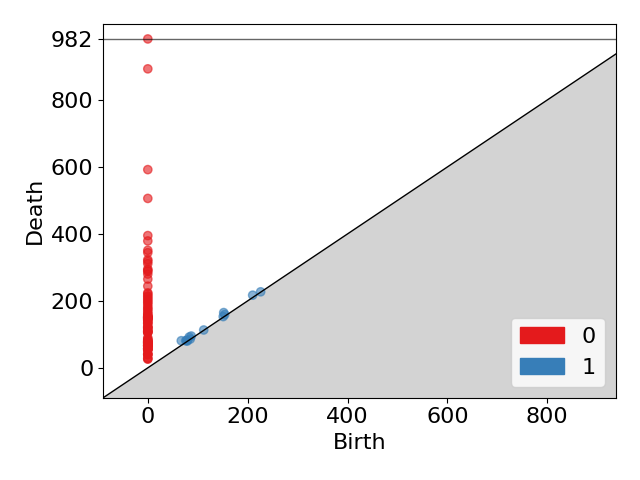}
&\includegraphics[width=0.2\textwidth]{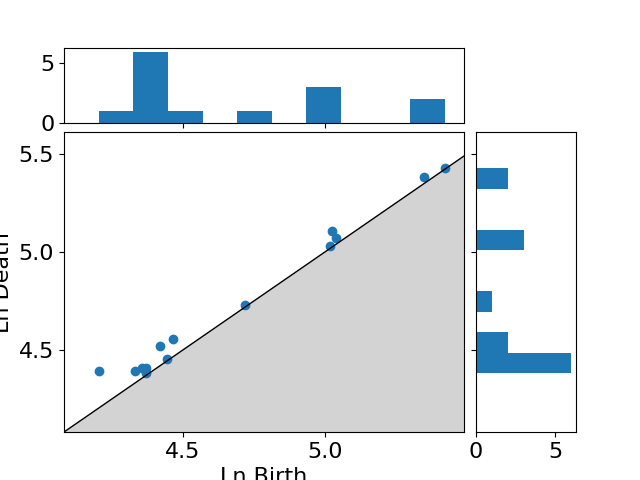}\\
1
&\includegraphics[width=0.2\textwidth]{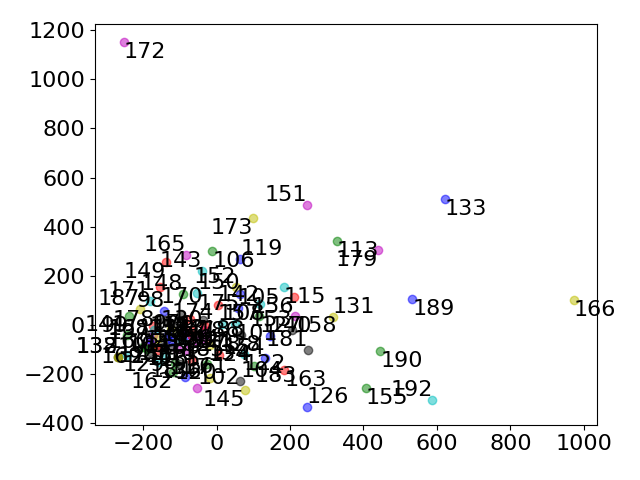}
%&\includegraphics[width=0.2\textwidth]{fig/stress1_tinyllama_0.9.png}
&\includegraphics[width=0.2\textwidth]{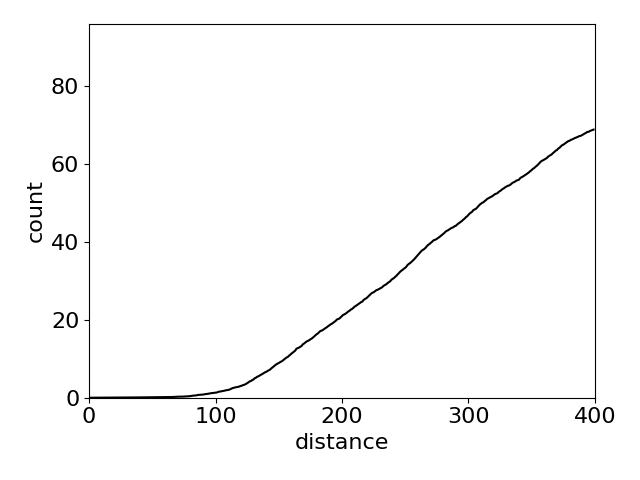}
%&\includegraphics[width=0.2\textwidth]{fig/barcode1_tinyllama_0.9.png}
&\includegraphics[width=0.2\textwidth]{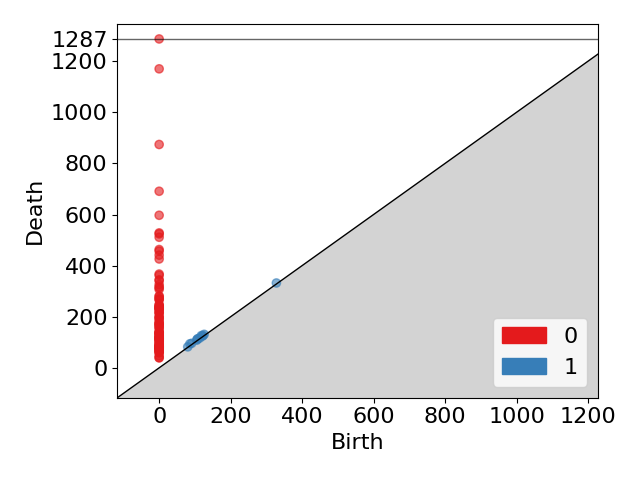}
&\includegraphics[width=0.2\textwidth]{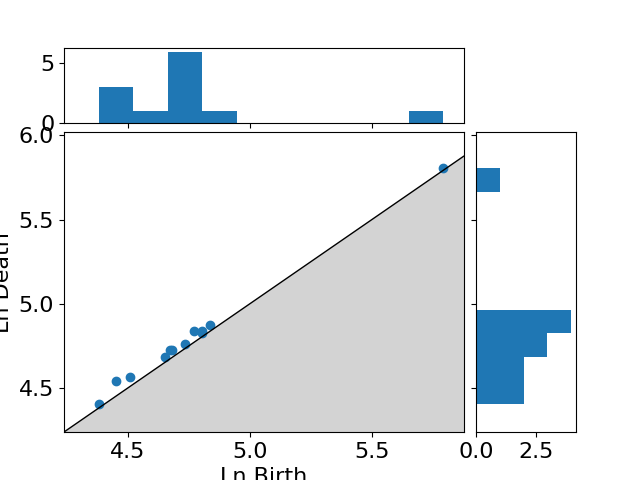}\\
2
&\includegraphics[width=0.2\textwidth]{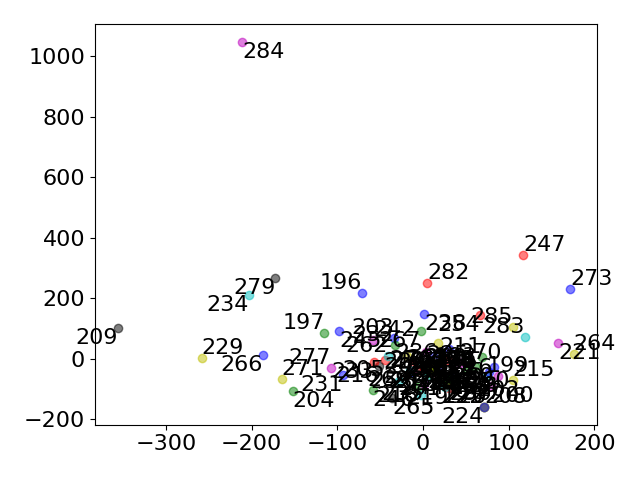}
%&\includegraphics[width=0.2\textwidth]{fig/stress2_tinyllama_0.9.png}
&\includegraphics[width=0.2\textwidth]{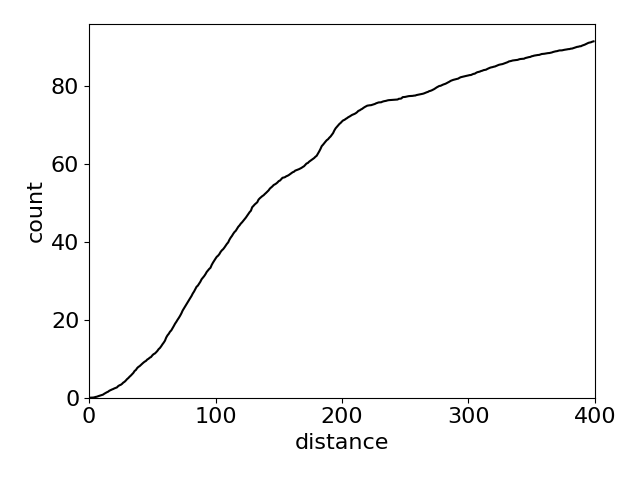}
%&\includegraphics[width=0.2\textwidth]{fig/barcode2_tinyllama_0.9.png}
&\includegraphics[width=0.2\textwidth]{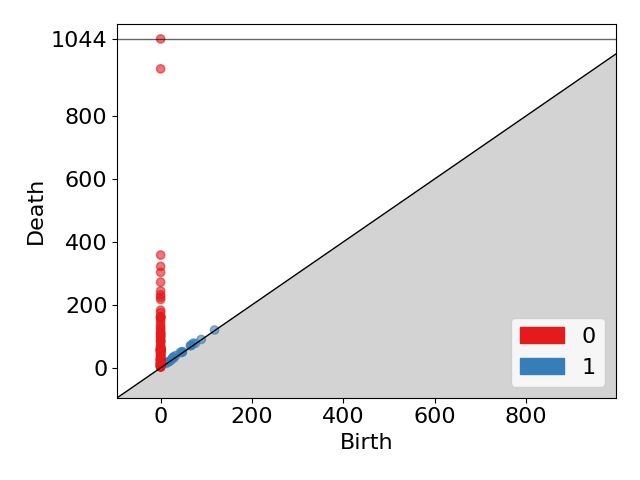}
&\includegraphics[width=0.2\textwidth]{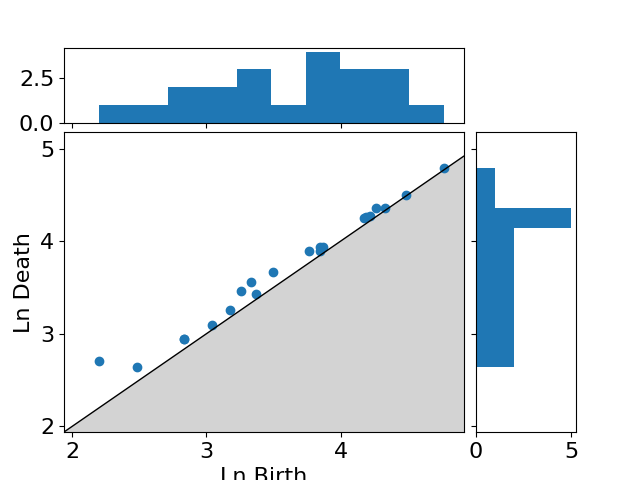}\\
3
&\includegraphics[width=0.2\textwidth]{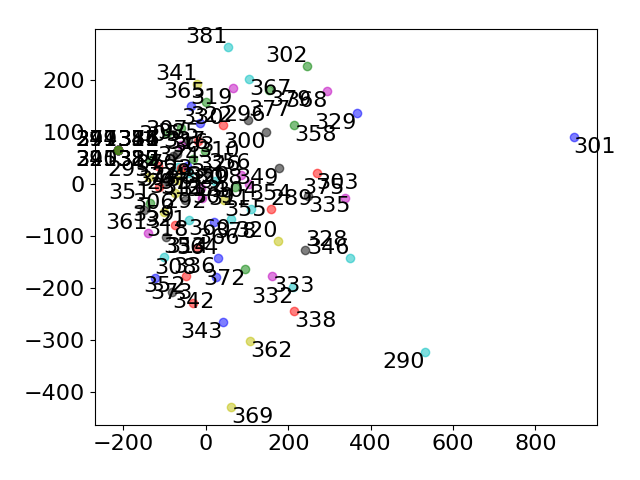}
%&\includegraphics[width=0.2\textwidth]{fig/stress3_tinyllama_0.9.png}
&\includegraphics[width=0.2\textwidth]{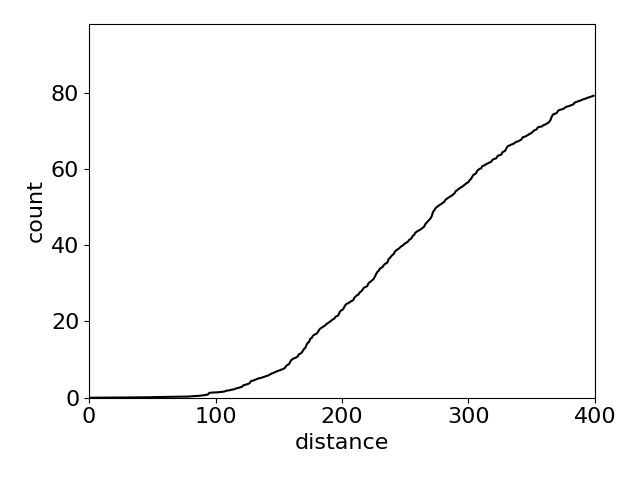}
%&\includegraphics[width=0.2\textwidth]{fig/barcode3_tinyllama_0.9.png}
&\includegraphics[width=0.2\textwidth]{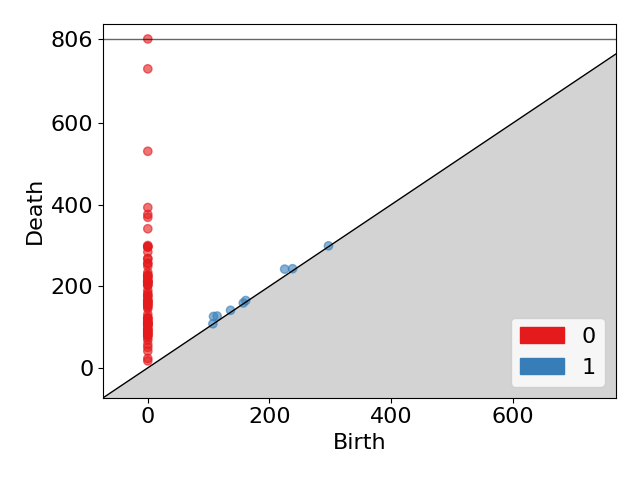}
&\includegraphics[width=0.2\textwidth]{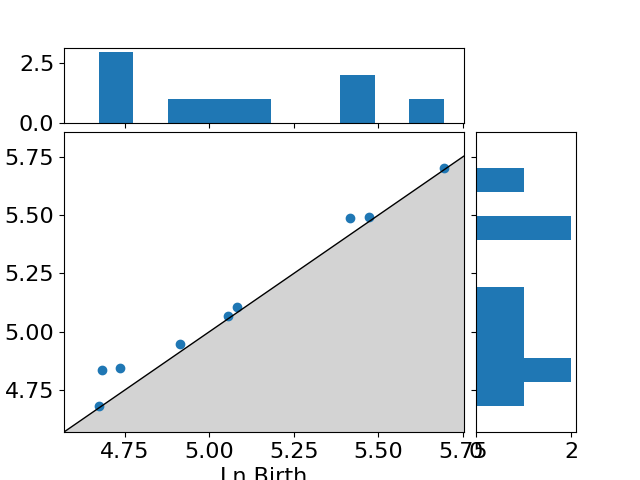}\\
4
&\includegraphics[width=0.2\textwidth]{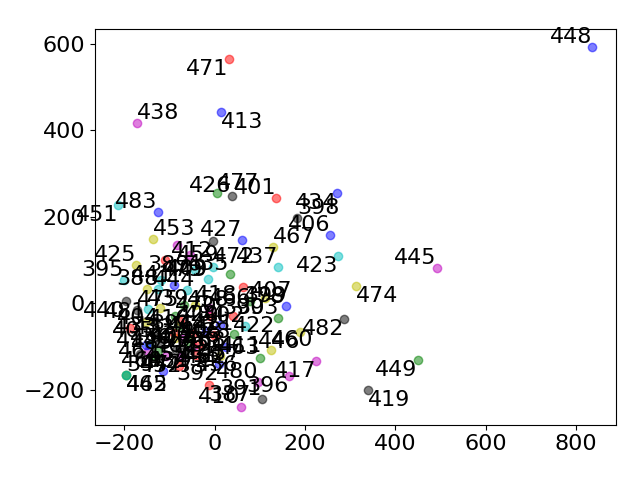}
%&\includegraphics[width=0.2\textwidth]{fig/stress4_tinyllama_0.9.png}
&\includegraphics[width=0.2\textwidth]{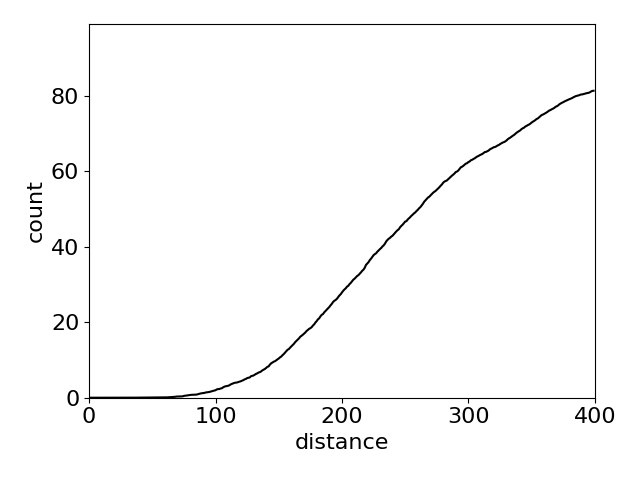}
%&\includegraphics[width=0.2\textwidth]{fig/barcode4_tinyllama_0.9.png}
&\includegraphics[width=0.2\textwidth]{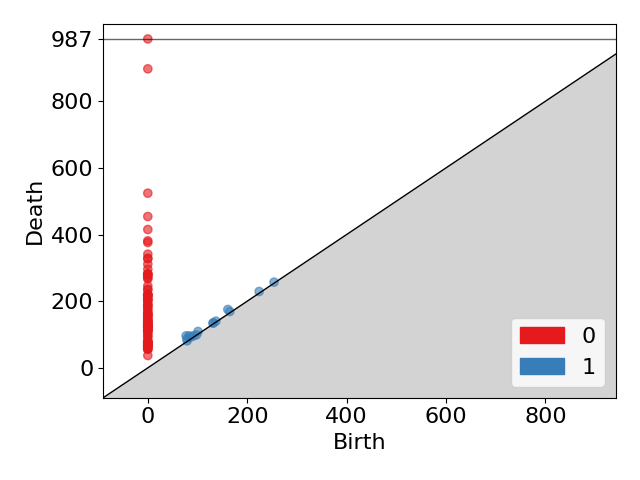}
&\includegraphics[width=0.2\textwidth]{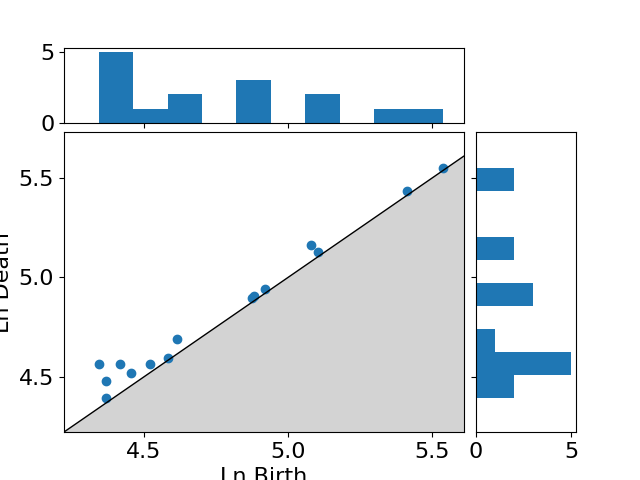}\\
5
&\includegraphics[width=0.2\textwidth]{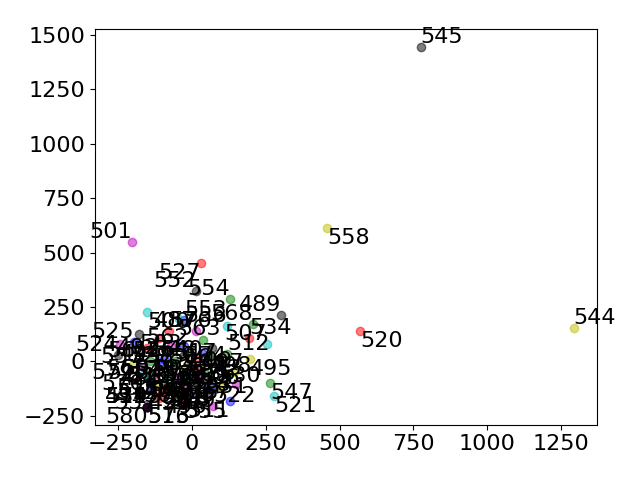}
%&\includegraphics[width=0.2\textwidth]{fig/stress5_tinyllama_0.9.png}
&\includegraphics[width=0.2\textwidth]{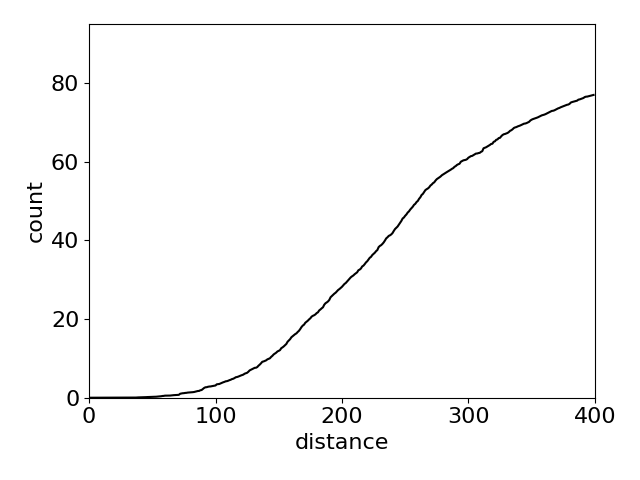}
%&\includegraphics[width=0.2\textwidth]{fig/barcode5_tinyllama_0.9.png}
&\includegraphics[width=0.2\textwidth]{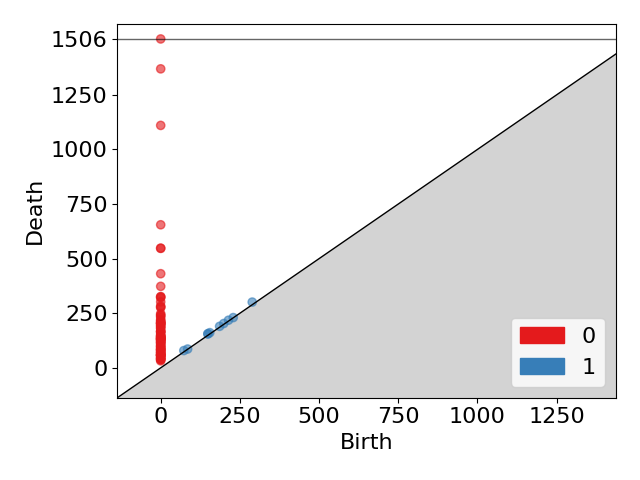}
&\includegraphics[width=0.2\textwidth]{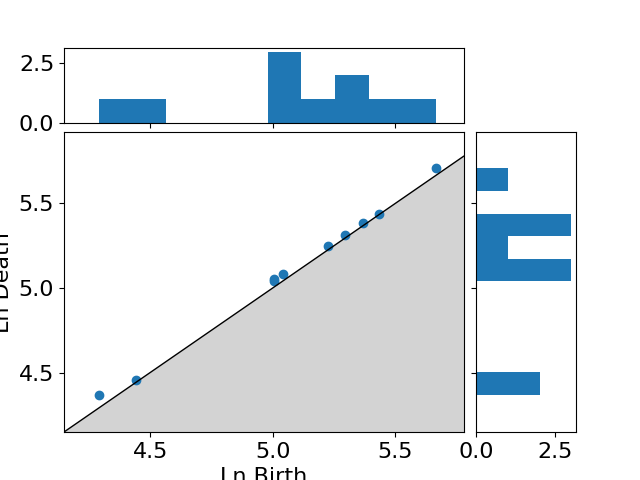}\\
6
&\includegraphics[width=0.2\textwidth]{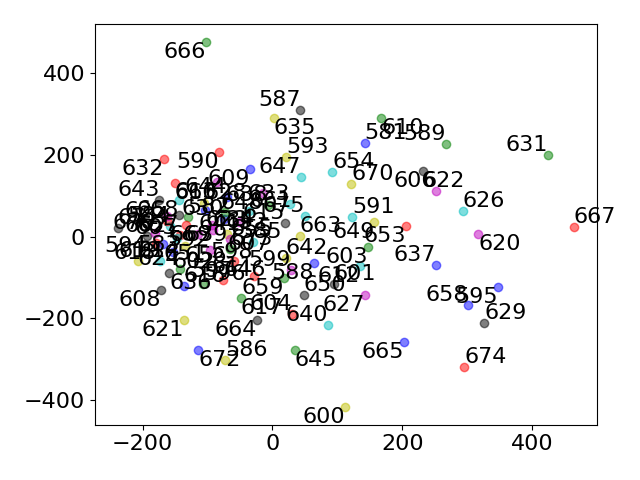}
%&\includegraphics[width=0.2\textwidth]{fig/stress6_tinyllama_0.9.png}
&\includegraphics[width=0.2\textwidth]{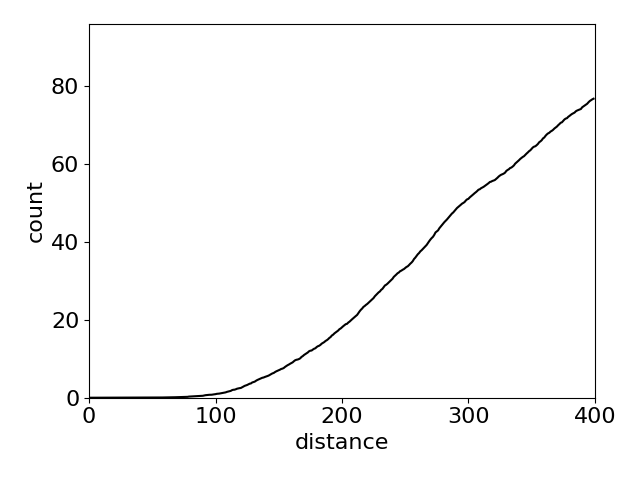}
%&\includegraphics[width=0.2\textwidth]{fig/barcode6_tinyllama_0.9.png}
&\includegraphics[width=0.2\textwidth]{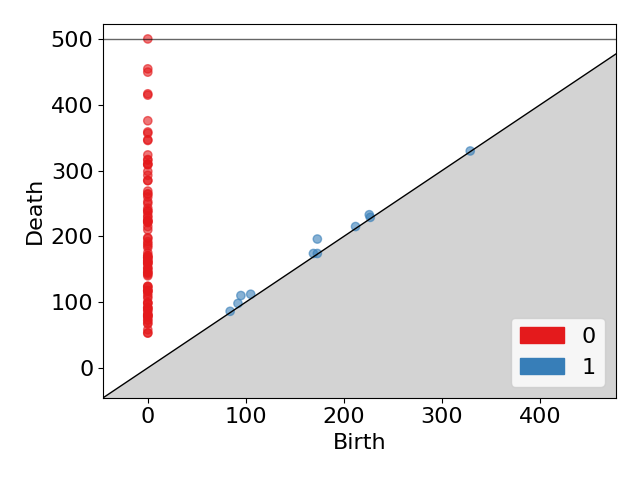}
&\includegraphics[width=0.2\textwidth]{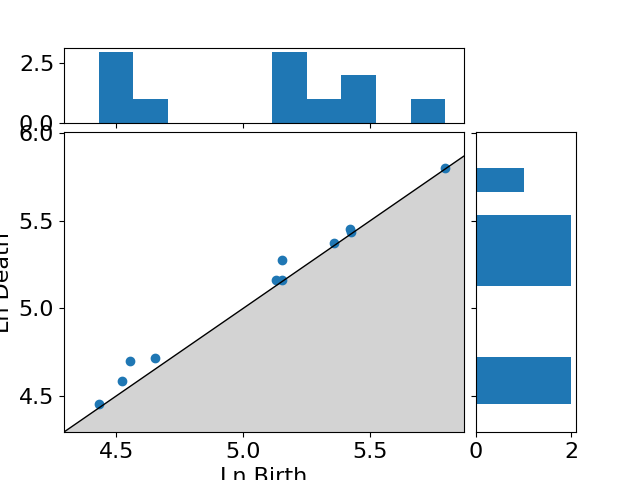}\\
\end{tabular}
\caption{Analysis of the program part of the output of TinyLlama with $t=0.9$}
\label{fig:tinyllama09}
\end{figure*}
 
\printbibliography %Prints bibliography

\end{document}